\documentclass[10pt,journal,compsoc]{IEEEtran}
\usepackage{amsmath,amsfonts}

\usepackage{array}
\usepackage{textcomp}
\usepackage{stfloats}
\usepackage{url}
\usepackage{verbatim}
\usepackage{graphicx}
\usepackage{cite}
\usepackage{multirow}
\usepackage{booktabs}

\usepackage{color}
\usepackage{hyperref}
\usepackage{graphicx}
\usepackage{amsmath}
\usepackage{amssymb}
\usepackage{cleveref}
\usepackage[section]{placeins}
\usepackage{float}
\usepackage{diagbox}

\usepackage{epstopdf}
\usepackage{amsmath}
\usepackage{amssymb}
\usepackage{multirow}
\usepackage{hyperref}

\usepackage{xcolor}

\usepackage{graphicx}%
\usepackage{multirow}%
\usepackage{amsmath,amssymb,amsfonts}%
\usepackage{amsthm}%
\usepackage{mathrsfs}%
\usepackage{xcolor}%
\usepackage{textcomp}%
\usepackage{manyfoot}%
\usepackage{booktabs}%
\usepackage{algorithm}%
\usepackage{algorithmicx}%
\usepackage{algpseudocode}%
\usepackage{listings}%

\usepackage{array}
\usepackage[caption=false,font=scriptsize,labelfont=sf,textfont=sf]{subfig}
\usepackage{stfloats}
\usepackage{url}
\usepackage{verbatim}
\usepackage{color}
\usepackage{hyperref}
\usepackage{cleveref}
\usepackage[section]{placeins}
\usepackage{float}
\usepackage{diagbox}
\usepackage{epstopdf}
\usepackage [latin1]{inputenc}
\usepackage{bbding}
\usepackage{booktabs}
\usepackage{makecell}
\usepackage{tabularx}
\usepackage{tikz}
\usetikzlibrary{shapes.geometric, arrows}

\ifCLASSINFOpdf
\else
\fi
\begin{document}

%
\title{Vision Mamba:\\ A Comprehensive Survey and Taxonomy}
%
%
%
%

\author{Xiao~Liu, Chenxu~Zhang, Lei~Zhang, \textit{Senior Member, IEEE}
        
\IEEEcompsocitemizethanks{\IEEEcompsocthanksitem This work was partially supported by National Key R\&D Program of China (2021YFB3100800), National Natural Science Fund of China (62271090), Chongqing Natural Science Fund (cstc2021jcyj-jqX0023), and National Youth Talent Project.  (Corresponding author: Lei Zhang) 

\IEEEcompsocthanksitem Xiao Liu, Chenxu Zhang and Lei Zhang are with the School of Microelectronics and Communication Engineering, Chongqing University, Chongqing 400044, China, and Peng Cheng Lab, Shenzhen, China.
(E-mail: liuxiao@stu.cqu.edu.cn, zhangchenxu@cqu.edu.cn, leizhang@cqu.edu.cn)

%


}

\thanks{Manuscript received April 19, 2005; revised August 26, 2015.}}

%
%

\markboth{Journal of \LaTeX\ Class Files,~Vol.~14, No.~8, August~2015}%
{Shell \MakeLowercase{\textit{et al.}}: Bare Demo of IEEEtran.cls for Computer Society Journals}
%



\IEEEtitleabstractindextext{%
\begin{abstract}
State Space Model (SSM) is a mathematical model used to describe and analyze the behavior of dynamic systems. This model has witnessed numerous applications in several fields, including control theory, signal processing, economics and machine learning. In the field of deep learning, state space models are used to process sequence data, such as time series analysis, natural language processing (NLP) and video understanding. By mapping sequence data to state space, long-term dependencies in the data can be better captured. In particular,  modern SSMs have shown strong representational capabilities in NLP, especially in long sequence modeling, while maintaining linear time complexity. Notably, based on the latest state-space models, Mamba \cite{Mamba} merges time-varying parameters into SSMs and formulates a hardware-aware algorithm for efficient training and inference. Given its impressive efficiency and strong long-range dependency modeling capability, Mamba is expected to become a new AI architecture that may outperform Transformer. Recently, a number of works have attempted to study the potential of Mamba in various fields, such as general vision, multi-modal, medical image analysis and remote sensing image analysis, by extending Mamba from natural language domain to visual domain. To fully understand Mamba in the visual domain, we conduct a comprehensive survey and present a taxonomy study. This survey focuses on Mamba's application to a variety of visual tasks and data types, and discusses its predecessors, recent advances and far-reaching impact on a wide range of domains. Since Mamba is now on an upward trend, please actively notice us if you have new findings, and new progress on Mamba will be included in this survey in a timely manner and updated on the Mamba project: \href{https://github.com/lx6c78/Vision-Mamba-A-Comprehensive-Survey-and-Taxonomy}{https://github.com/lx6c78/Vision-Mamba-A-Comprehensive-Survey-and-Taxonomy}.
\end{abstract}

}

\maketitle

\IEEEdisplaynontitleabstractindextext

%
\IEEEpeerreviewmaketitle

\IEEEraisesectionheading{\section{Introduction}\label{sec1}}
\IEEEPARstart{S}{ince} the emergence of deep learning, convolutional neural networks (CNNs) and Transformers have become prevalent in various visual domain tasks. CNNs have gained popularity due to their simple architecture and scalability \cite{ImageNet}, \cite{CNN_Residual}, \cite{EfficientNet}. However, the introduction of Vision Transformer (ViT)\cite{vit} has disrupted the landscape. This can be attributed to the self-attention mechanism that enables global sensory fields and dynamic weights. The emergence of foundation models has brought changes and opportunities to AI in recent years. In particular, foundation models based on Transformer architecture with self-attention mechanism have an extensive range of applications and exciting performance. However, CNNs struggle to long-range dependencies, while Transformers bear the burden of secondary computational complexity.

Recently, State Space Model (SSM) has shown significant effectiveness of state space transformations in capturing the dynamics and dependencies of language sequences. \cite{S4} introduces a Structured State Space Sequence Model (S4) specifically designed to model remote dependencies with the advantage of linear complexity, bringing a new impetus to the development of natural language understanding. The success of a new variant, formally referred to as Mamba \cite{Mamba}, has been proven to be capable of higher performance on actual data with up to a million length sequences. This has made it a hot topic recently. Mamba combines selective scanning (S6) and enjoys fast inference with linear scaling of sequence lengths. Its computational advantage is more than five times faster than Transformer.

Since natural language and computer vision have gradually formed a trend of integration, extending Mamba from large language model (LLM) with outstanding performance to computer vision is an attractive direction. Vim \cite{VisionME} is the first pure SSM-based model to handle intensive prediction tasks and the first application of SSM to a generic backbone in vision. The framework addresses two challenges of Mamba in processing image sequences: modeling unidirectionality and lack of positional awareness. VMamba \cite{VMamba} proposes a cross-scan strategy to bridge the gap between 1D array and 2D sequence scanning. Mamba-ND \cite{Mamba-ND} aims to extend the Mamba architecture to a variety of vision tasks on multidimensional data, including 1D, 2D and 3D data, and more other vision tasks such as image restoration \cite{MambaIR}, \cite{Activating_S_Resolution}, \cite{CU-Mamba}, \cite{VmambaIR}, infrared small target detection \cite{MiM-ISTD}, point clouds \cite{PointMamba}, \cite{PCM}, \cite{Point_Mamba}, \cite{3DMambaComplete}, and video modeling \cite{VideoMamba}, \cite{Video_M_S}, \cite{RhythmMamba}. These works explore the potential of Mamba and show promise in the field of vision.

\begin{table*}[htbp]
\caption{A Taxonomy of Vision Mamba.}\label{Mamba_Category}%
\setlength\tabcolsep{0.5pt}
\belowrulesep=0pt
\aboverulesep=0pt

\begin{tabular*}{\textwidth}{@{\extracolsep\fill}lccc}
\toprule
Category & Sub-category & Method & Details \\
\midrule
\multirow{17}{*}{General Vision}
    & High/Mid-level Vision
         & \makecell[l]{\textbf{Backbone: }Vim \cite{VisionME}, VMamba \cite{VMamba}, Mamba-ND \cite{Mamba-ND}, \\
            LocalMamba \cite{LocalMamba}, EfficientVMamba \cite{EfficientVMamba}, SiMBA \cite{SiMBA}, \\ 
            PlainMamba \cite{PlainMamba}, $[V]$-Mamba \cite{[V]-Mamba}, DGMamba \cite{DGMamba} \\
            \textbf{Video Analysis and Understanding: }VideoMamba \cite{VideoMamba}, \\
            Video Mamba Suite \cite{Video_M_S}, RhythmMamba \cite{RhythmMamba} \\
            \textbf{Vertical-domain Vision: }Res-VMamba \cite{Res-VMamba}, \\
            InsectMamba \cite{InsectMamba}, MambaAD \cite{MambaAD}, MiM-ISTD \cite{MiM-ISTD}, \\
            MemoryMamba \cite{MemoryMamba} \\
            }
         & \makecell[cp{3.5cm}]{\textbf{data type:} Image, Video\\
         \textbf{highlight:} Scanning strategy, Architectural optimization, Transfer learning, Domain generalization} \\
    \cline{2-4}
    & Low-level Vision
        & \makecell[l]{\textbf{Image Denoising: }UMV-Net \cite{U-shaped-Mamba-Dehazing}, FreqMamba \cite{FreqMamba}\\
        \textbf{Image Restoration: }MambaIR \cite{MambaIR}, MMA \cite{Activating_S_Resolution}\\
        CU-Mamba \cite{CU-Mamba}, VmambaIR \cite{VmambaIR}, Retinexmamba \cite{Retinexmamba} \\}
        & \makecell[cp{3.5cm}]{\textbf{data type:} Image\\
        \textbf{highlight:} Hybrid CNN-Mamba models, Frequency domain analysis, Attention mechanisms, Scanning direction}\\
    \cline{2-4}
    & 3-D Visual Recognition
        & \makecell[l]{\textbf{Point Could Analysis: }PointMamba \cite{PointMamba}, PCM \cite{PCM}, \\
        Point Mamba \cite{Point_Mamba}, 3DMambaComplete \cite{3DMambaComplete}, \\
        \textbf{Hyperspectral Imaging Analysis: }Mamba-FETrack \cite{Mamba-FETrack}}
        & \makecell[cp{3.5cm}]{\textbf{data type: } Point cloud, hyperspectral imaging\\
        \textbf{highlight:} Reordering strategy, Z-order, Octree-based ordering strateg, HyperPoint Generation, Spectral dimension analysis} \\
    \cline{2-4}
    & Visual Generation
    & \makecell[l]{ZigMa \cite{ZigMa}, Motion Mamba \cite{Motion_Mamba}, Gamba \cite{Gamba}, Matten \cite{Matten}, \\
    SMCD \cite{SMCD}\\}
        & \makecell[cp{3.5cm}]{\textbf{data type:} Image, Temporal sequences\\
        \textbf{highlight:} Diffusion models, Attention mechanisms, Scanning direction, Gaussian splatting}\\

\midrule
\multirow{7}{*}{Multi-Modal}
    & Heterologous Stream
         & \makecell[l]{\textbf{Multi-Modal Understanding: }MambaTalk \cite{MambaTalk}, ReMamber \cite{ReMamber},\\ SpikeMba \cite{SpikeMba} \\
            \textbf{Multimodal large language models: }VL-Mamba \cite{VL-Mamba}, Cobra \cite{Cobra}\\
            }
         & \makecell[cp{3.5cm}]{\textbf{data type:} Text, Image, Speech, Video\\
         \textbf{highlight:} Gesture Synthesis, Large language models, Video grounding, Multi-Modal Understanding} \\
    \cline{2-4}
    & Homologous Stream
        & \makecell[l]{Sigma \cite{Sigma}, Fusion-Mamba \cite{Fusion-Mamba-Detection}}
        & \makecell[cp{3.5cm}]{\textbf{data type:} infrared image, X-modality, RGB image\\
        \textbf{highlight:} feature fusion, Mamba's gate mechanism, Channel-Attention operation}\\
\midrule

\multirow{25}{*}{Vertical Application}
    & Remote Sensing Image
         & \makecell[l]{
         \textbf{Remote Sensing Image Processing: }Pan-Mamba \cite{Pan-Mamba}, \\
         HSIDMamba \cite{HSIDMamba}\\
         \textbf{Remote Sensing Image Classification: }RSMamba \cite{RSMamba}, \\
         SpectralMamba \cite{SpectralMamba}, SS-Mamba \cite{SS-Mamba}, S2Mamba \cite{S2Mamba}\\
         \textbf{Remote Sensing Image Change Detection: }ChangeMamba \cite{ChangeMamba}, \\
         RSCama \cite{RSCama}\\
         \textbf{Remote Sensing Image Segmentation: }Samba \cite{Samba}, \\
         RS3Mamba \cite{RS3Mamba}, RS-Mamba \cite{RS-Mamba}\\
         \textbf{Remote Sensing Image Fusion: }FusionMamba \cite{FusionMamba_Efficient}, LE-
Mamba \cite{State_Sharing_Fusion}\\
            }
         & \makecell[cp{3.5cm}]{\textbf{data type:} Remote sensing images \\
         \textbf{highlight:} Pan-sharpening, Position embedding, Hybrid Mamba-MLP models, Self-attention, Information fusion, Frequency domain analysis, Large language models, Selective scan} \\
    \cline{2-4}
    & Medical Image
        & \makecell[l]{\textbf{Medical Image Segmentation: }U-Mamba \cite{U_Mamba}, VM-UNet \cite{VM-UNet}, \\
         Mamba-UNet \cite{Mamba-UNet}, LightM-UNet \cite{LightM-UNet}, LMa-UNet \cite{LMa-UNet}, \\ VM-UNET-V2 \cite{VM-UNET-V2},
         Mamba HUNet \cite{Mamba-HUNet}, TM-UNet \cite{TM-UNet}, \\
         Swin-UMamba \cite{Swin-UMamba}, P-Mamba \cite{P-Mamba}, H-vmunet \cite{H-vmunet}, \\
         Semi-Mamba-UNet \cite{Semi-Mamba-UNet}, Weak-Mamba-UNet \cite{Weak-Mamba-UNet}, \\
         UltraLight VM-UNet \cite{UltraLight_VM-UNet}, ProMamba \cite{ProMamba}, \\
         SegMamba \cite{SegMamba}, nnMamba \cite{nnMamba}, T-Mamba \cite{T-Mamba}, Vivim \cite{Vivim} \\
            \textbf{Pathological Diagnosis: }MedMamba \cite{MedMamba}, MamMIL \cite{MamMIL}, \\CMViM \cite{CMViM}, \\
            MambaMIL \cite{MambaMIL}, SurvMamba \cite{SurvMamba} \\
            \textbf{Deformable Image Registration: }MambaMorph \cite{MambaMorph}, \\VMambaMorph \cite{VMambaMorph} \\
            \textbf{Medical Image Reconstruction: }FDVM-Net \cite{FDVM-Net}, MambaMIR \cite{MambaMIR}, \\
            FusionMamba \cite{FusionMamba_Dynamic}, MambaDFuse \cite{MambaDFuse} \\
            \textbf{Other Medical Tasks: }MD-Dose \cite{MD-Dose}, MMH \cite{Dual-Camera_Tracker} \\
            }
         & \makecell[cp{3.5cm}]{\textbf{data type:} \textit{CT: }\cite{U_Mamba},\cite{VM-UNet},\cite{LightM-UNet},\cite{LMa-UNet},\cite{SegMamba},\cite{nnMamba},\cite{T-Mamba},\cite{MambaMorph},\cite{FusionMamba_Dynamic},\cite{MambaDFuse} \\\textit{MRI:}\cite{U_Mamba},\cite{Swin-UMamba},\cite{Mamba-UNet},\cite{LMa-UNet},\cite{Mamba-HUNet},\cite{Semi-Mamba-UNet},\cite{Weak-Mamba-UNet},\cite{SegMamba},\cite{nnMamba},\cite{CMViM},\cite{MambaMorph},\cite{FusionMamba_Dynamic},\cite{MambaDFuse}\\
         \textit{skin lesion:}\cite{VM-UNet},\cite{VM-UNET-V2},\cite{H-vmunet},\cite{TM-UNet},\cite{UltraLight_VM-UNet} \\
         \textit{Endoscopy images:}\cite{Swin-UMamba},\cite{VM-UNET-V2},\cite{H-vmunet},\cite{TM-UNet},\cite{ProMamba},\cite{MedMamba},\cite{FDVM-Net},\cite{Dual-Camera_Tracker} \\\textit{X-ray images:}\cite{LightM-UNet},\cite{MedMamba}\\
         \textit{Echocardiogram:}\cite{P-Mamba},\cite{MedMamba}\\
         \textit{Video:}\cite{Vivim}\\
         \textit{Whole slide images:}\cite{MamMIL},\cite{MambaMIL},\cite{SurvMamba}\\
         \textit{dose distribution maps:}\cite{MD-Dose} \\
         \textbf{highlight:} U-shaped architecture, Hierarchical architecture, Lightweight, Hybrid CNN-Mamba models, pure Mamba models, Prompt, Weakly-supervised learning, Sequence reordering, Attention mechanism, Information fusion} \\
    \cline{2-4}

\bottomrule
\end{tabular*}
\end{table*}

Extensive work has been done on various aspects of the Mamba architecture to explore the applications of SSM for vertical-domain visual tasks. Specifically, owing to Mamba's exceptional computational efficiency and long-range dependency modeling capability, numerous vertical-domain Mambas have been rapidly emerged for medical image analysis. Since U-net \cite{U-Net} is a popular network architecture in medical image segmentation, U-Mamba \cite{U_Mamba} was the first to combine U-net with Mamba as a hybrid CNN-SSM model to process high-resolution medical image data. This approach outperforms state-of-the-art CNNs and Transformer-based segmentation networks in terms of efficiency and performance. Following this, a number of Mamba variants for medical image analysis have been proposed. 
Additionally, the community has proposed several extensions to explore the boundaries of Mamba's capabilities in remote sensing image processing, analysis and understanding \cite{Pan-Mamba}, \cite{RSMamba}, \cite{Samba}, \cite{RS3Mamba}, \cite{RS-Mamba}, \cite{ChangeMamba}, \cite{FusionMamba_Efficient}, \cite{SpectralMamba}, \cite{State_Sharing_Fusion}, \cite{HSIDMamba}. 

Our survey stands out from existing ones on Mamba \cite{SSM_survey}, \cite{Visual_Mamba_survey}, \cite{Mamba-360}, \cite{Vision_Mamba_survey}, showcasing unique strengths. First, we employ a clear and precise taxonomy of Mamba variants, systematically organizing and summarizing Mamba's progress and applications in computer vision and its vertical domains. Compared to current reviews, our taxonomy is more targeted and reader-friendly, facilitating easier understanding and navigation of relevant content for researchers.
Second, we provide insightful and rational taxonomy of each subcategory, ensuring comprehensive coverage of different vertical domains and issues. This taxonomy not only enhances readers' understanding of the characteristics and application scenarios of each category but also assists researchers in finding the information they need in specific areas.
Third, we ensure thorough explanations of the principles and technical details of each method, aiming to deepen readers' understanding of their intentions. Our review article not only offers conceptual explanations but also provides concrete examples and experimental results to facilitate readers' comprehension and application of these methods.
Fourth, we include two additional sections in the survey, namely "data type" and "highlight" which provide supplementary categorization and explanations for the methods within each category. These additional sections offer researchers more references and options, helping them quickly locate and understand relevant content.

Overall, this work presents a survey and taxonomy of state space models (SSMs) in vision oriented methodologies and applications, aiming to help researchers understand the latest advances related to Mamba modeling, while discarding the fragmented work of SSM in language field. 
To further discuss the progress of Mamba based on the latest SSM, we categorize the Mamba models according to the downstream application scenarios or tasks, as shown in Table \ref{Mamba_Category}. The main categories include general vision tasks, multi-modal tasks and vertical-domain tasks, where vertical-domain tasks contain remote sensing image analysis and medical image analysis. General vision tasks are comprised of  high-level/mid-level vision, low-level vision, 3D vision and Multiple Types of Data Streams. Since it is difficult to distinguish between Mamba in high-level and mid-level vision for the objectives, we lump them together into one category. Typical tasks of low-level vision include image processing, image restoration and image generation. Mamba for 3D vision mainly refers to point cloud analysis for 3-D visual recognition. Mamba's strong global modeling capabilities and linear computational complexity make it well-suited for handling the intrinsic irregularity and sparsity of point clouds. Given that multi-modal tasks have recently become a popular direction, an introduction of Mamba in multi-modal tasks is presented. Compared to general vision tasks, remote sensing scenarios are more complex and diverse. The variable spatial-temporal resolution in CNN-based and Transformer-based approaches presents challenges for modeling accuracy and memory usage. Therefore, Mamba has attracted researchers' attention and effort to achieve a commendable balance of performance and efficiency in remote sensing image analysis tasks. Another widely-studied category of Mamba is medical image analysis, which is well-suited to adapt Mamba in conjunction with other architectures or methodologies in handling high-resolution inputs and detailed information.

The rest of the article is structured as follows. Section 2 discusses the basic architecture and principle of SSMs. Section 3 summarises the specific aspects of Mamba in general vision domain. Section 4 describes the application of Mamba in vision-lanugage multi-modal learning tasks. Additionally, Section 5 analyses the architecture and methodology of Mamba in vertical domains including remote sensing and medical image analysis. Finally, Section 6 concludes this survey, discuss several challenges and give some promising directions for Mamba research and application. 

\section{Formulation of Mamba}

\begin{figure}[t]
\begin{center}
\includegraphics[width=1\linewidth,height=0.5\linewidth]{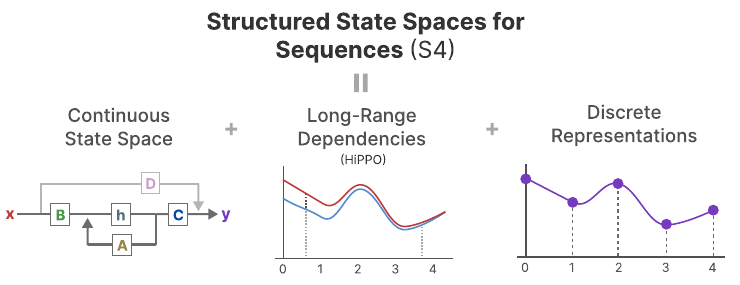}
\end{center}
   \caption{S4 \cite{S4} of the structural block diagram. It consists of three parts: state space models, HiPPO for handling long-range dependencies and discretization for creating recurrent and convolution representations.}
\label{S4}
\end{figure}

State Space Models (SSMs)  \cite{S4}, \cite{Continuous_time_ssm}, \cite{Diagonal_SSM}, \cite{Long_Sequence_SSM}, \cite{Resurrecting_RNLS}, \cite{Mamba} use an intermediate state variable to achieve sequence-to-sequence mapping, allowing for handling long sequences. Structured State Space-based models, such as S4 \cite{S4}, use low-rank corrections to regulate some of the model's parameters, enabling stable diagonalization and reducing the SSMs to a well-studied Cauchy kernel. This approach solves the problem of excessive computation and memory requirements of previous SSM models. S4 is a significant improvement over previous sequence modeling approaches. Mamba \cite{Mamba} improves SSMs by introducing a selection mechanism and hardware-aware algorithm that parameterizes SSMs based on the input sequence. This solves the discrete modality issue and achieves more straightforward processing when dealing with long sequences in language and genomics.

\subsection{State Space Models}
The primary representation of SSMs is the continuous time representation, as illustrated in Fig.~\ref{S4}, which is defined by four parameters ($\boldsymbol A$, $\boldsymbol B$, $\boldsymbol C$, $\boldsymbol D$). This representation transforms a time-dependent set of inputs $x(t)\in \mathbb{R}^{}$ into a set of outputs $y(t)\in \mathbb{R}^{}$ using a hidden state $h(t)\in \mathbb{R}^{N}$. The mapping process can be expressed as follows:
\begin{equation}
\begin{aligned}
\label{state_output_equation} {h'(t)} &= \boldsymbol{A} h(t) + \boldsymbol{B} x(t),\\
{y(t)} &= \boldsymbol{C} h(t) + \boldsymbol{D} x(t)
\end{aligned}
\end{equation}

Specifically, the first equation in Eq.~\ref{state_output_equation}, i.e., the state equation, is operated by multiplying the matrix $\boldsymbol B$ with the input $x(t)$ followed by the matrix $\boldsymbol A$ with the previous state $h(t)$. Parameter $\boldsymbol A$ is the evolution parameter, which stores all the previous history information represented by a matrix of coefficients and determines the degree of influence of the previous hidden state on the hidden state used to update the spatial state of the hidden state at the next moment of time in $\boldsymbol A$. The projection parameter matrix $\boldsymbol B$  that determines how much the input $x(t)$ affects the hidden state.
The second equation in Eq.~\ref{state_output_equation}, i.e., the output equation, describes how the hidden state is transformed into output via the projection parameter matrix  $\boldsymbol C$  and how the input affects the output via the matrix $\boldsymbol D$. In general, the parameter  $\boldsymbol D$ provides a direct signal from the input to the output, also known as a skip-connection, and a simplification can be made by omitting  $\boldsymbol D$, i.e., assuming  $\boldsymbol D = 0$.

As shown in Fig.~\ref{S4}, S4 is inspired by the continuous system used for 1D sequences and uses sample timescale parameters $\Delta$ to transform from continuous SSM to discrete SSM. Thus, the mapping from continuous signal input $x(t)$ to output $y(t)$ is transformed into a sequence-to-sequence mapping: $x_k \rightarrow y_k$, which facilitates the processing of discrete token input, such as images and texts.

There are two stages in this sequence-to-sequence process. The first stage is discretization, which serves as the principled foundation of heuristic gating mechanisms. The continuous parameters $(\boldsymbol A, \boldsymbol B)$ are converted to discrete parameters $(\boldsymbol{\overline{A}}, \boldsymbol{\overline{B}})$ using sample timescale parameters $\Delta$. This conversion is usually done using zero-order hold (ZOH), resulting in discrete formulas:
\begin{equation}
\begin{aligned}
\label{state_output_discretization} {\boldsymbol{\overline{A}}} &= e^{\Delta \boldsymbol A},\\
{\boldsymbol{\overline{B}}} &= (\Delta\boldsymbol {A})^{-1}(e^{\Delta \boldsymbol A} - I) \cdot \Delta\boldsymbol{B},\\
{h_k} &= {\boldsymbol{\overline{A}}} h_{k-1} + {\boldsymbol{\overline{B}}} x_{k},\\
{y_k} &= {\boldsymbol{C}} h_{k-1}\\
\end{aligned}
\end{equation}
where $\boldsymbol{\overline{A}} \in \mathbb{R}^{N \times N}$, $\boldsymbol{\overline{B}} \in \mathbb{R}^{N \times 1}$ and  $\boldsymbol{{C}} \in \mathbb{R}^{1 \times N}$.

The second stage is computation, where the model computes the output by global convolution:

\begin{equation}
\begin{aligned}
\label{state_output_global_convolution}
{\boldsymbol{\overline{K}}} &= ({\boldsymbol{C}}{\boldsymbol{\overline{B}}}, {\boldsymbol{C}}{\boldsymbol{\overline{AB}}}, \ldots, {\boldsymbol{C}}{\boldsymbol{\overline{A}^{L-1}}}\boldsymbol{\overline{B}}),  \\
{y} &= {x} * \boldsymbol{\overline{K}}\\
\end{aligned}
\end{equation}
where $L$ is the length of the input sequence $x$, and ${\boldsymbol{\overline{K}}} \in \mathbb{R}^{L}$ is a structured convolutional kernel.

Since three of the discrete parameters $\boldsymbol A$, $\boldsymbol B$ and $\boldsymbol C$ are constants when the inputs are constants and the convolutional kernel $\boldsymbol K$ is static, the process of recomputing the inputs to states and states to outputs is avoided, and it can be realized to train in parallel like a convolutional neural network for efficient computation. In order to fully utilize the respective advantages of the convolution and recurrent modes, SSMs employ parallel training using convolution mode in the training phase and efficient autoregressive inference using the recurrent mode. This is because generating the output of the next time step requires only the state of the current time step rather than the entire input history. However, in recurrent mode, the recurrent neural networks (RNNs) tends to forget a certain information over time.

To solve this problem, Linear State Space Layer (LSSL) \cite{Continuous_time_ssm} uses HiPPO Continuous Time Memory Theory \cite{HiPPO}, which attempts to compress all currently observed input signals into a vector of coefficients. It uses matrix $\boldsymbol A$ to build a state representation that captures recent tokens well and decays older tokens. The $\boldsymbol A$ matrix constructed by HiPPO contains $N$ polynomial compressors, used to compress the continuously growing historical data and adjust the specific values of the coefficients during training process. However, the continuously increasing order causes dimensionality explosion, so the authors transform the hidden states obtained from the state equations in Eq.~\ref{state_output_equation} into outputs through linear combination, i.e., using the projection parameter matrix $\boldsymbol C$ of the output equations.

Furthermore, to address the complexity issue arising from the discrete-time SSM due to the repeated matrix multiplications with $\boldsymbol A$, S4 proposes a practical solution. It enhances matrix $\boldsymbol A$ by leveraging a structural result to simplify the SSMs. Specifically, the author employs the Normal Plus Low-Rank (NPLR) approach to HiPPO matrix decomposition. This approach allows for stable diagonalization and significantly simplifies the computation of a Cauchy kernel, making the method more practical and applicable in real-world scenarios.

\subsection{Selective State Space Models (S6)}

\begin{figure*}[t]
\begin{center}
\includegraphics[width=1\linewidth,height=0.35\linewidth]{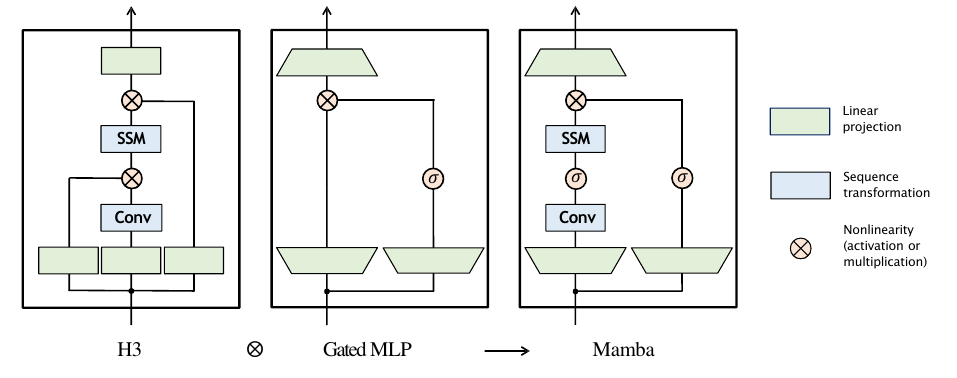}
\end{center}
   \caption{Architecture of Mamba block (image from \cite{Mamba})}
\label{mamba}
\end{figure*}

\textbf{1) Selective Mechanism.} The previous linear time-invariant state-space model, due to the lack of content awareness, cannot achieve selective tasks based on input content, which leads S4 to spend equal attention on all tokens. However, in reality, the importance of tokens is different, and the degree of importance changes dynamically with the training process. Therefore, spending more effort on the important content and dynamically adjusting the importance level to match the complex input content is more effective.

Based on the above considerations, Mamba merges the selectivity mechanism into the state space model to obtain Selective State Space Models (S6). Specifically, in order to operate on an input sequence with batch size $B$, length $L$ and $D$ channels, Mamba aims to apply SSM independently for each channel.
In Mamba~\cite{Mamba}, the matrices $\boldsymbol B$, matrix $\boldsymbol C$, and $\Delta$ in S4 are functions of the inputs, allowing the model to adaptively adjust its behavior according to the inputs. The discretization process after incorporating the selection mechanism into the model is as follows:
\begin{equation}
\begin{aligned}
\label{state_output_S6} {\boldsymbol{\overline{B}}} &= s_B(x),\\
{\boldsymbol{\overline{C}}} &= s_C(x),\\
{\Delta} &= \tau _A(Parameter + s_A(x))\\
\end{aligned}
\end{equation}
where $\boldsymbol{\overline{B}} \in \mathbb{R}^{B \times L \times N}$, $\boldsymbol{\overline{C}} \in \mathbb{R}^{B \times L \times N}$ and $\Delta \in \mathbb{R}^{B \times L \times D}$. $s_B(x)$ and $s_C(x)$ are linear functions that project the input $x$ into a $N$-dimensional space, while $s_A(x)$ projects the hidden state dimension $D$ linearly into the desired dimension, connected to the RNN gating mechanism. Through the above computations, the parameters $\Delta$, $\boldsymbol B$, $\boldsymbol C$ become input functions with length $L$, transforming the time-invariant model into a time-varying model, thus achieving selectivity.

The size of $\Delta$ has been changed from $D$ to $(B, L, D)$, meaning that for each token in a batch (there are a total of $B \times L$), there is a unique $\Delta$ for input data dependency and more fine-grained control functions. The larger the step size of $\Delta$, the more the model focuses on the inputs, rather than the stored state. Instead, the smaller the step size, the more the model will ignore the specific inputs, and thus focus more on the stored state. 
Parameters $\boldsymbol B$ and $\boldsymbol C$ become input data dependent on the function of S4, thus allowing finer control over whether input $x$ goes to state $h$ or state $h$ goes to output $y$.
Parameter  $\boldsymbol A$ does not become data dependent, but after the discretization operation of SSM, it is possible to make $\boldsymbol A$ relevant to the input through the data dependency of $\Delta$. Meanwhile, since the parameter $\boldsymbol A$ has dimension $N$, it has different roles in each SSM dimension, thus achieving an accurate generalization of all the previous contents rather than a simple compression.

\textbf{2) Hardware-Aware State Expansion.} The selectivity mechanism overcomes the limitations of previous linear time-invariant state-space models. However, the time-varying nature poses a computational challenge. In Mamba, since the input and output at this time are not simply static mapping relationships, efficient convolutional computation with fixed convolutional kernels cannot be used. So, the researchers designed a hardware-aware parallel algorithm in cyclic mode. The model is computed by scanning instead of convolution. Mamba's scanning algorithm avoids the pitfalls of RNNs that cannot be performed in parallel. Since the new state in the scan operation needs previous state to be computed, parallel scanning cannot be achieved by using loop computation directly. Researchers found that each state is actually the whole compression of the previous states, that is, the previous states can be directly used to compute the new state. Then it can be assumed that the order of the execution of the operation is independent of the associated attributes. Therefore, Mamba implements the selective scan algorithm by calculating sequences in segments and combining them iteratively, together with the parameters on which the input data depends.

On the other hand, since GPUs have many processors and can perform highly parallel computations, Mamba takes advantage of the GPU's HBM and fast SRAM IOs to avoid frequent SRAM writes to the HBM by using kernel fusion. Specifically, Mamba performs discretization and recursive operations in the higher-speed SRAM memory and then writes the output back to the HBM. When inputs are loaded from the HBM to the SRAM, the intermediate state is not preserved but is recomputed in backpropagation.

As shown in Fig.~\ref{mamba}, Mamba combines the base blocks of SSM with the MLP blocks prevalent in modern neural networks to form a new Mamba block, which is stacked and combined with normalization and residual connection to form the Mamba network architecture.

\subsection{Discussion and Summary}
The selective mechanism enables the Mamba to possess linear computational complexity and long-range dependencies modeling capabilities, and the hardware-aware state expansion makes it memory efficient. With these two key techniques, Mamba shows great potential in various applications beyond previous state space models.

\section{Mamba in General Vision Tasks}
\begin{figure*}[t]
\centering
\subfloat[BiDirectional Scan \cite{VisionME}]{
		\includegraphics[scale=0.65]{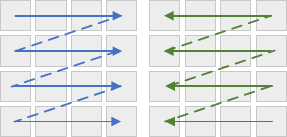}} \hspace{2mm}
\subfloat[Cross Scan \cite{VMamba}]{
		\includegraphics[scale=0.65]{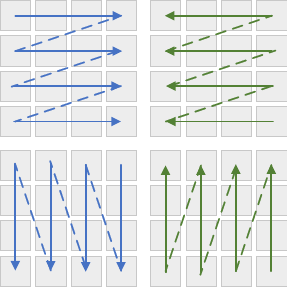}}  \hspace{2mm}
\subfloat[Continuous 2D Scanning \cite{PlainMamba}]{
		\includegraphics[scale=0.65]{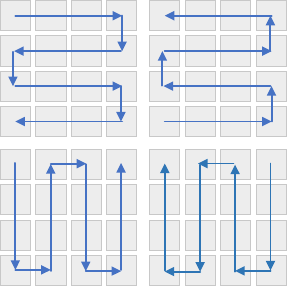}}  \hspace{2mm}
\subfloat[Hilbert scanning \cite{MambaAD}]{
		\includegraphics[scale=0.65]{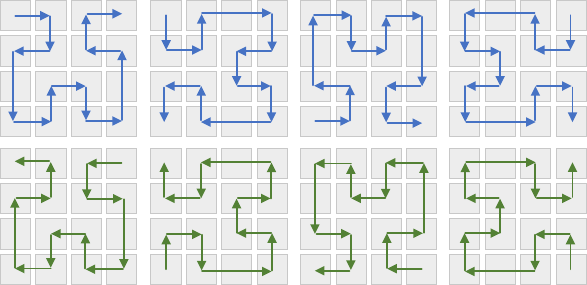}}
\\
\subfloat[Local Scan \cite{LocalMamba}]{
		\includegraphics[scale=0.65]{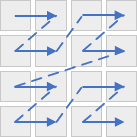}}  \hspace{2mm}
\subfloat[Efficient 2D Scanning \cite{EfficientVMamba}]{
		\includegraphics[scale=0.65]{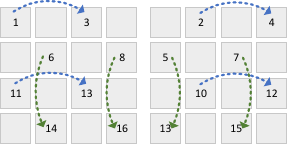}}   \hspace{2mm}
\subfloat[Spatiotemporal Selective Scan \cite{VideoMamba}]{
		\includegraphics[scale=0.65]{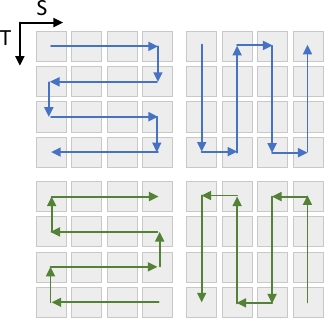}}  \hspace{2mm}
\subfloat[Zigzag Scan \cite{ZigMa}]{
		\includegraphics[scale=0.65]{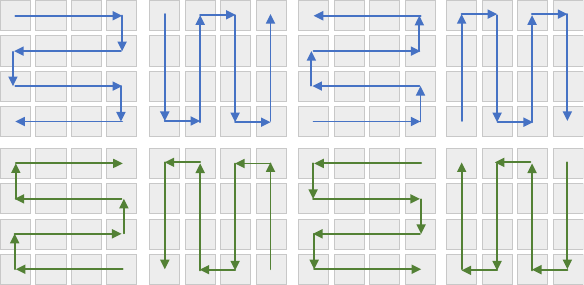}}
\caption{Comparison of different selective scan methods in Vim, VMamba, Plain Mamba, MambaAD, LocalMamba, EfficientVMamba, VideoMamba and ZigMa.}
\label{scan}
\end{figure*}

This section reviews the application of Mamba and its variants to general vision, including high-level/mid-level Vision, Low-level Vision and 3D vision. In the following subsections, we introduce Mamba variants redesigned for each task. Considering the importance of selective scanning strategies in vision tasks, we further summarise the existing classical 2D scanning mechanisms in Fig.~\ref{scan}.

\subsection{Mamba for High-level/Mid-level Vision}

\subsubsection{Vision Backbone with Mamba}
The success in language modeling has motivated researchers to design generic and efficient visual backbones using advanced Mamba model. 

\begin{figure*}[t]
\begin{center}
\includegraphics[width=1\linewidth,height=0.27\linewidth]{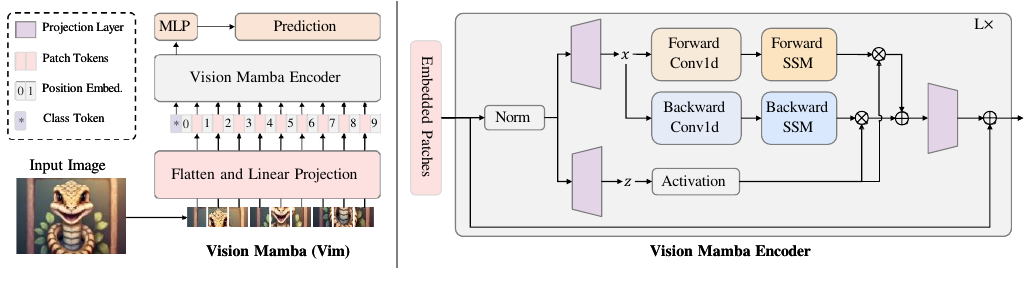}
\end{center}
   \caption{An overview of the first vision Mamba framework (Vim) (image from \cite{VisionME}).}
\label{Vim}
\end{figure*}

Vim \cite{VisionME} presents the first pure SSM-based model to handle intensive prediction tasks. The authors claim that SSM has 
two challenges in vision application: modeling unidirectionality and lack of location awareness. For this reason, Vim employs the technique of bidirectional SSM and positional embedding. As shown in Fig.~\ref{Vim}, to handle the vision task, it first transforms the multi-dimensional image ${t} \in \mathbb{R}^{H \times W \times C}$ into a spread two-dimensional block $x_p \in \mathbb{R}^{J \times (P^2 \cdot C )}$, where ($H$, $W$) is the size of the input image, $C$ is the number of channels and $P$ is the size of the image block. Similar to the Transformer's position embedding approach, Vim linearly projects $x_p$ to a vector of size $D$ and adds position embedding $E_{pos} \in \mathbb{R}^{(J+1) \times D}$, which provides a sense of the spatial information and also uses class token (CLS) to represent the entire patch sequence. Finally, the token sequence is fed into layer $l$ of the Vim encoder, and the output $T$ is obtained. In contrast to the normal Mamba block, Vim uses a bidirectional SSM block, where the inputs of the SSM block are processed from the forward and backward directions, respectively. The outputs of the forward and backward processes are computed through SSM. Then, the outputs y of the two parts are selected by the gating signal $z$ and added together to obtain the output token sequence $T$. In Vim \cite{VisionME}, experiments are conducted on ImageNet image classification, COCO object detection, and ADE20k semantic segmentation tasks, respectively. The results show that Vim outperforms the highly optimized ViT variant at different scales. Vim also outperforms the traditional ResNet network and DeiT in terms of performance, computational efficiency and memory consumption in a variety of visual tasks, indicating that Vim has a great potential for high-resolution downstream visual applications and long-sequence multi-modal applications.

\begin{figure}[t]
\begin{center}
\includegraphics[width=0.65\linewidth,height=1\linewidth]{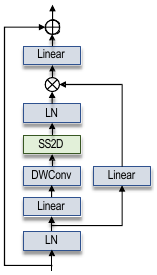}
\end{center}
   \caption{Visual state space (VSS) block (image from \cite{VMamba}).}
\label{vmamba}
\end{figure}

Due to the non-causal nature of visual data, directly applying Mamba to patches and flat images will inevitably lead to a restricted receptive field, as the relationship to unscanned patches cannot be estimated. VMamba \cite{VMamba} refers to this problem as direction-sensitive. In this case, the approach needs to consider the spatial structure and global relevance of the data rather than focusing only on the order of data. As shown in  Fig.~\ref{scan}(b), a Cross Scan Module (CSM) is designed to traverse the spatial domain and convert any non-causal visual image into a sequence of sequential patches using a four-way scanning strategy. Specifically, image blocks are chosen to be expanded into sequences along rows and columns, and then scanned in four different directions: from left to bottom right, from bottom right to top left, from top right to bottom left, and from bottom left to top right. Then, each sequence is reshaped into a single image, and all the sequences are combined into a new sequence, i.e., scanned in a sequence from the four corners of the feature map to the opposite position to integrate the information from all other pixels in different directions. VMamba chooses to integrate S6 with CSM owing to its linear complexity of the selective scanning, and preserves the global receptive field as a core element in the construction of a visual state space (VSS) block, called SS2D, as illustrated Fig.~\ref{vmamba}. Besides, since some convolutional structures in Mamba naturally take into account local spatial relations between pixels, VMamba uses a hierarchical structure, unlike Vim, and does not use position embedding bias.

The overall flow of VMamba~\cite{VMamba} is as follows:

\begin{itemize}
	\item Divide the input image into patches using a ViT-like Stem module, which generates a feature map of dimension: ${\frac{H}4} \times {\frac{W}4} \times C_1$.
	\item Divide the whole network into four stages, stacking several VSS blocks in each stage and keeping the same dimensionality.
	\item Downsample the features from the previous stage by a patch merger operation and used as input features for the next stage.
	\item Repeat this process to create separate features with different resolutions for each of the four different scales produced.
\end{itemize}

Experimental results show that VMamba outperforms established benchmarks and performs well in vision tasks. All model-size variants of VMamba outperform all competing pairs, including ResNet, DeiT, Swin and ConvNeXt. VMamba also outperforms Vim on the three basic tasks of classification, detection and semantic segmentation.

Pei et al. \cite{EfficientVMamba} propose Efficient VMamba model, which employs an atrous-based selective scanning strategy to achieve a lightweight model design. As shown in  Fig.~\ref{scan}(f), the authors proposed Efficient 2D Scanning (ES2D) to reduce the complexity of scanning by skipping the sampling of each patch on the feature map. This procedure decomposes the global scanning method into local and global sparse forms. Skip sampling of local receptive fields reduces computational complexity and improves feature extraction efficiency by selectively scanning smaller blocks of the feature map. Global information extraction is achieved by combining the processed patches to reorganize the feature map. Between each ES2D block, additional convolutional branches are introduced as a complement to the original global SSM branches, which contains channel attention module i.e., SE \cite{SE} to adjusts the trade-off between local and global information. More importantly, compared to Vim and VMamba, the authors observe from the results that the pipeline in this paper significantly reduces complexity and suggests that hybrid models of SSM, convolution and fusion operations may benefit from extensive detailed information for more diverse features. 

Similarly, LocalMamba \cite{LocalMamba} shows that improvements in scanning algorithms are vital to improving the performance of vision mamba. Therefore, a windowed selective scan is proposed, as shown in  Fig.~\ref{scan}(e). The image is divided into several different local windows, where independent scans are performed to maintain the dependence of the original neighboring tokens. In addition to the windowed selective scan, the authors also consider horizontal and vertical scans. However, directly using different scanning methods in each layer will significantly increase the computational requirements. To solve this problem, the authors define the scanning direction searched in each layer as a search space based on the principle of DARTS \cite{DARTS}. 

PlainMamba~\cite{PlainMamba} further explores the improvement of selective scanning algorithms in non-hierarchical networks. As shown in Fig.~\ref{scan}(c), the model uses a continuous 2D scanning process to ensure the spatial adjacency of tokens. More importantly, the direction-aware updating technique is proposed to encode direction information to distinguish the token for the next scanning.

Finding a generic design for extending the Mamba architecture to arbitrary multi-dimensional data is also a meaningful problem. In order to accommodate inputs of different dimensions, Mamba-ND \cite{Mamba-ND} employs some bidirectional or multi-directional scanning order design. Given there are many ways to order multidimensional data, Mamba-ND performs SSM calculations only along the forward or backward direction of various dimensional axes. Then it uses the above mentioned method to arrange these Mamba layers. Numerous experiments have shown that uses the standard 1D-SSM layer and an alternating directional scanning is not only simple but also superior to more complex designs.

In addition to changes directly in the selective scanning approach, generic visual models combined with other methodologies are equally attractive. Patro et al. \cite{SiMBA} point out that Mamba is unstable when scaled up to large networks. Therefore, SiMBA is proposed that a Mamba structure similar to a hierarchical Transformer, i.e., a combination of a Mamba block and a multi-layer perceptron (MLP). This study bridges the performance gap in small-scale networks by using a nonlinear channel combination approach with MLP. Further, the model combines Mamba with EinFFT to solve the stability problems in both small and large networks. It is worth noting that the application of learnable Fourier transformation aims to comply with the conditions for the stability of linear state space models.

In \cite{[V]-Mamba} and \cite{DGMamba}, the potential of generic Mamba models for transfer learning is explored. [V]-Mamba \cite{[V]-Mamba} exploits the migration mechanisms of Linear Probing (LP) and Visual Prompting (VP) methods. By comparing with the classical Transformer model ViT, the model reveals a weak positive correlation between model size and performance on both LP and VP approaches, respectively. DGMamba \cite{DGMamba} explores the generalization of Mamba to unseen domains. It avoids the corruption of domain invariant feature learning by reducing the non-semantic information in the hidden state. In other words, the authors argue that such non-semantic information is accumulated and amplified with propagation, which hampers the generalization performance of the model. The process is represented as follows: 
\begin{equation}
\begin{aligned}
\label{DGMamba} {y_t} &= \boldsymbol{\overline{C}}h_t,\\
{\boldsymbol{\overline{C}}} &= \boldsymbol{C} \times M,\\
M &= ({\Delta \boldsymbol{A}} \textgreater \alpha) + (1 - ({\Delta \boldsymbol{A}} \textgreater \alpha)){\Delta \boldsymbol{A}}.\\
\end{aligned}
\end{equation}
where $\alpha$ is the confidence threshold and the coefficient parameters for $\Delta \boldsymbol{A} \leq \alpha$ will be suppressed. 

For the samples with the highest confidence percentage, DGMamba replaces background patches of input exhibiting low Grad-CAM scores \cite{Grad-CAM} with counterparts from diverse domains. This process reduces the overfitting of the model to simple samples. Then for the remaining samples, it applies Prior-Free Scanning to randomly shuffle the background patches. The average generalization performance of DGMamba on the PACS dataset is 2.7\% higher than the SOTA method, which achieves the SOTA performance on the Office-Home dataset in all scenarios. On VLCS dataset, DGMamba exhibits excellent average generalization performance compared to the SOTA method.

We summarise the results of these models in Table \ref{Mamba_ImageNet}, Table \ref{Mamba_COCO} and Table \ref{Mamba_ADE20k} for comparisons in terms of backbone network development.

\begin{table*}[t]
\caption{Comparison Between Vision Backbones with Mamba on ImageNet-1K \cite{ImageNet} datasets for Image Classification. $^{\dag}$ represents Vim model is fine-tuned with long sequence setting. $^{*}$ represents LocalVim model without scan direction search.}\label{Mamba_ImageNet}%
\setlength\tabcolsep{0.5pt}

\begin{tabular*}{\textwidth}{@{\extracolsep\fill}lcccc}
\toprule
Method & Image size & \#Params (M). & FLOPs (G) & Top-1 ACC (\%). \\
\toprule
\textbf{CNN} \\
\toprule
ResNet-101 \cite{ResNet} & $224^2$ & 45 & - & 77.4 \\
ResNet-152 \cite{ResNet} & $224^2$ & 39 & - & 78.3 \\
RefNetY-4G \cite{RefNetY} & $224^2$ & 21 & 4.0 & 80.0 \\
RefNetY-8G \cite{RefNetY} & $224^2$ & 39 & 8.0 & 81.7 \\
RefNetY-16G \cite{RefNetY} & $224^2$ & 84 & 16.0 & 82.9 \\
ConvNet-T & $224^2$ & 29 & 4.5 & 82.1 \\
\toprule
\textbf{Transformer} \\
\toprule
ViT-B/16 \cite{vit} & $384^2$ & 86 & 55.4 & 77.9 \\
ViT-L/16 \cite{vit} & $384^2$ & 307 & 190.7 & 76.5 \\
DieT-S \cite{DieT} & $224^2$ & 22 & 4.6 & 79.8 \\
DieT-B \cite{DieT} & $224^2$ & 86 & 17.5 & 81.8 \\
DieT-B \cite{DieT} & $384^2$ & 86 & 55.4 & 83.1 \\
Swin-T \cite{Swin} & $224^2$ & 29 & 4.5 & 81.3 \\
Swin-S \cite{Swin}  & $224^2$ & 50 & 8.7 & 83.0 \\
Swin-B \cite{Swin}  & $224^2$ & 88 & 15.4 & 83.5 \\
\toprule
\textbf{Mamba}\\
\toprule
Vim-Ti \cite{VisionME} & $224^2$ & 7 & - & 73.1 \\
Vim-Ti$^{\dag}$ \cite{VisionME} & $224^2$ & 7 & - & 78.3 \\
Vim-S \cite{VisionME} & $224^2$ & 26 & - & 80.5 \\
Vim-S$^{\dag}$ \cite{VisionME} & $224^2$ & 26 & - & 81.6 \\
\midrule
VMamba-T \cite{VMamba} & $224^2$ & 22 & 4.5 & 82.2 \\
VMamba-S \cite{VMamba} & $224^2$ & 44 & 9.1 & 83.5 \\
VMamba-B \cite{VMamba} & $224^2$ & 75 & 15.2 & 83.2 \\
\midrule
Mamba-2D-S \cite{Mamba-ND} & $224^2$ & 24 & - & 81.7 \\
Mamba-2D-B \cite{Mamba-ND} & $224^2$ & 92 & - & 83.0 \\
\midrule
LocalVim-T$^{*}$ \cite{LocalMamba} & $224^2$ & 8 & 1.5 & 75.8 \\
LocalVim-T \cite{LocalMamba} & $224^2$ & 8 & 1.5 & 76.2 \\
LocalVim-S$^{*}$ \cite{LocalMamba} & $224^2$ & 28 & 4.8 & 81.0 \\
LocalVim-S \cite{LocalMamba} & $224^2$ & 8 & 4.8 & 81.2 \\
LocalVMamba-T \cite{LocalMamba} & $224^2$ & 26 & 5.7 & 82.7 \\
LocalVMamba-S \cite{LocalMamba} & $224^2$ & 50 & 11.4 & 83.7 \\
\midrule
EfficientVMamba-T \cite{EfficientVMamba} & $224^2$ & 6 & 0.8 & 76.5 \\
EfficientVMamba-S \cite{EfficientVMamba} & $224^2$ & 11 & 1.3 & 78.7 \\
EfficientVMamba-B \cite{EfficientVMamba} & $224^2$ & 33 & 4.0 & 81.8 \\
\midrule
SiMBA-S(Monarch) \cite{SiMBA} & $224^2$ & 18.5 & 3.6 & 81.1 \\
SiMBA-S(EinFFT) \cite{SiMBA} & $224^2$ & 15.3 & 2.4 & 81.7 \\
SiMBA-S(MLP) \cite{SiMBA} & $224^2$ & 26.5 & 5.0 & 84.0 \\
SiMBA-B(Monarch) \cite{SiMBA} & $224^2$ & 26.9 & 5.5 & 82.6 \\
SiMBA-B(EinFFT) \cite{SiMBA} & $224^2$ & 22.8 & 4.2 & 83.0 \\
SiMBA-B(MLP) \cite{SiMBA} & $224^2$ & 40.0 & 9.0 & 84.7 \\
SiMBA-L(Monarch) \cite{SiMBA} & $224^2$ & 42 & 8.7 & 83.8 \\
SiMBA-L(EinFFT) \cite{SiMBA} & $224^2$ & 36.6 & 7.6 & 83.9 \\
SiMBA-L(MLP)$^{\dag}$ \cite{SiMBA} & $224^2$ & 66.6 & 16.3 & 49.4 \\
\midrule
PlainMamba-L1 \cite{PlainMamba} & $224^2$ & 7 & 3.0 & 77.9 \\
PlainMamba-L2 \cite{PlainMamba} & $224^2$ & 25 & 8.1 & 81.6 \\
PlainMamba-L3 \cite{PlainMamba} & $224^2$ & 50 & 14.4 & 82.3 \\

\bottomrule
\end{tabular*}
\end{table*}

\subsubsection{Video Analysis and Understanding}

Mamba's ability to process long sequences is ideally suited for video analysis and understanding tasks, where complete contextual information needs to be mined. Its advantages in efficiency and performance trigger Mamba to several video based vision tasks, such as video understanding \cite{VideoMamba}, \cite{Video_M_S} and remote physiological estimation \cite{RhythmMamba}.

VideoMamba \cite{VideoMamba} introduces self-distillation into the non-hierarchical Vim architecture. It aims to address the dual challenges of local redundancy and global dependency in video understanding. Initially, the spatiotemporal scanning applies bi-directional Mamba block layers to the temporal and spatial dimensions. Then different scanning variants extend the original 2D scan to a various bi-directional 3D scan, as shown in Fig.~\ref{scan}(g). Experimentally, it is found that the spatial-first-based bi-directional scanning is optimal by comparing to temporal-first and spatiotemporal scanning strategies. By evaluating on ImageNet-1K, compared to the recent Mamba work such as Vim and VMamba and the traditional Transformer-based approach such as TimeSformer \cite{TimeSformer} and ViViT \cite{ViViT}, it has shown significant progress in processing speed and performance. 
Chen et al. \cite{Video_M_S} consider the Vim's forward SSM and backward SSM as two separated branches. The technique uses the inverse design and shares the weights of the two SSM branches, which allows the input features to capture joint spatiotemporal information. It is evaluated with the temporal model, multi-modal interaction network and spatial-temporal model on 12 video comprehension tasks. The experimental results showcases the strong potential of Mamba for both video-only and video-language tasks. 

Different from the previous Transformer and Mamba based methods in video understanding, RhythmMamba \cite{RhythmMamba} is designed to enhance the Mamba's capability of different length video. Firstly, the input features are segmented into fragments of different time lengths and computed separately using SSM. Secondly, the extraction of spatial information is achieved by using diffusion, self-attention and frame average pooling techniques before the Mamba block. In order to effectively focus on the periodic nature of the weak signal rPPG, the channel swapping is used in the frequency domain after SSM modeling of the features. Finally, the rPPG features are exploited for prediction.

\subsubsection{Vertical-domain Vision}
Mamba can also assist in many vertical-domain vision problems, such as food detection and classification, etc. This section focuses on vision Mamba's variants and applications in several specific scenarios.

Res-VMamba \cite{Res-VMamba} is a model specifically designed for food classification tasks. It uses a global residual mechanism in the first three stages of the VMamba structure. During training, Res-VMamba utilises both the local detail information obtained from each VSS block and the global information. 
In addition, \cite{InsectMamba} integrates SSM/MSA/CNN/MLP into Mamba to cope with the challenges of high artefacts and species diversity. It uses MLP and Softmax to obtain the weight factors with respect to various coded features featuring selective adaptive aggregation.
\cite{MambaAD} proposes a Hilbert scanning method mixed with other scanning methods in the Hybrid State Space Block instead of the plain Mamba block, which significantly improves the feature sequence modeling. Moreover, comprehensive experiments on six different anomaly detection datasets and seven evaluation metrics demonstrate the state-of-the-art performance.

Recently, MiM-ISTD \cite{MiM-ISTD} explores the possibility of Mamba in infrared small target detection (ISTD). Because Mamba is not good at capturing these critical local features, MiM-ISTD considers local patches as visual sentences and uses an outer layer of Mamba to explore global information. Then, each visual sentence is decomposed into sub-patches as visual words, and the inner Mamba is used between visual words. The encoder of MiM-ISTD is hierarchically structured with four stages, each consisting of multiple MiM blocks to process word-level and sentence-level features. Correspondingly, its decoder and encoder have the same number of stages in the resnet block composition. 

MemoryMamba \cite{MemoryMamba} is the first attempt for defect recognition in industrial applications, which designs Mem-SSM blocks with coarse- and fine-grained memory encoding to capture and utilize historical defect-related data efficiently.

\subsection{Mamba for Low-level Vision}

The potential of Mamba in low-level Vision tasks is presented and discussed, even in the primary stages. 
Such tasks are usually characterized by the fact that the input and output of the model are images.

\begin{table*}[t]
\caption{Comparison Between Vision Backbone with Mamba using Mask-RCNN detector on COCO Dataset \cite{COCO} for Object Detection and Instance Segmentation.}\label{Mamba_COCO}%
\setlength\tabcolsep{0.5pt}

\begin{tabular*}{\textwidth}{@{\extracolsep\fill}lcccccccc}
\toprule
Method & AP$^{b}$ & AP$^{b}_{50}$ & AP$^{b}_{75}$ & AP$^{m}$ & AP$^{m}_{50}$ & AP$^{m}_{75}$ & \#Params (M). & FLOPs (G) \\
\toprule
\textbf{CNN} \\
\toprule
ResNet-101 \cite{ResNet} & 38.2 & 58.8 & 41.4 & 34.7 & 55.7 & 37.2 & 44 & 260 \\
ConvNeXt-T \cite{ConvNeXt}  & 44.7 & 65.8 & 48.3 & 40.1 & 63.3 & 42.8 & 48 & 262 \\
ConvNeXt-S \cite{ConvNeXt}  & 45.4 & 67.9 & 50.0 & 41.8 & 65.2 & 45.1 & 70 & 400\\
\toprule
\textbf{Transformer} \\
\toprule
Swin-T \cite{Swin} & 42.7 & 65.2 & 46.8 & 39.3 & 62.2 & 42.2 & 48 & 267 \\
Swin-S \cite{Swin} & 44.8 & 66.6 & 48.9 & 40.9 & 63.2 & 44.2 & 69 & 354\\
PVTv2-B3 \cite{PVTv2}  & 47.0 & 68.1 & 51.7 & 42.5 & 65.7 & 45.7 & 65 & 397 \\
\toprule
\textbf{Mamba}\\
\toprule
Vim-Ti \cite{VisionME} & 45.7 & 63.9 & 49.6 & 39.2 & 60.9 & 41.7 & - & - \\
\midrule
VMamba-T \cite{VMamba} & 46.5 & 68.5 & 50.7 & 42.1 & 65.5 & 45.3 & 42 & 262 \\
VMamba-S \cite{VMamba} & 48.2 & 69.7 & 52.5 & 43.0 & 66.6 & 46.4 & 64 & 357 \\
VMamba-B \cite{VMamba} & 48.5 & 69.6 & 53.0 & 43.1 & 67.0 & 46.4 & 96 & 482 \\
\midrule
LocalVMamba-T \cite{LocalMamba} & 46.7 & 68.7 & 50.8 & 42.2 & 65.7 & 45.5 & 45 & 291 \\
LocalVMamba-S \cite{LocalMamba} & 48.4 & 69.9 & 52.7 & 43.2 & 66.7 & 46.5 & 69 & 414 \\
\midrule
EfficientVMamba-T \cite{EfficientVMamba} & 37.5 & 57.8 & 39.6 & - & - & - & 13 & - \\
EfficientVMamba-S \cite{EfficientVMamba} & 39.1 & 60.3 & 41.2 & - & - & - & 19 & - \\
EfficientVMamba-B \cite{EfficientVMamba} & 42.8 & 63.9 & 45.8 & - & - & - & 44 & - \\
\midrule
SiMBA-S \cite{SiMBA} & 46.9 & 68.6 & 51.7 & 42.6 & 65.9 & 45.8 & 60 & 382 \\
\midrule
PlainMamba-Adapter-L1 \cite{PlainMamba} & 44.1 & 64.8 & 47.9 & 39.1 & 61.6 & 41.9 & 31 & 388 \\
PlainMamba-Adapter-L2 \cite{PlainMamba} & 46.0 & 66.9 & 50.1 & 40.6 & 63.8 & 43.6 & 53 & 542 \\
PlainMamba-Adapter-L3 \cite{PlainMamba} & 46.8 & 68.0 & 51.1 & 41.2 & 64.7 & 43.9 & 79 & 696 \\
\bottomrule
\end{tabular*}
\end{table*}

\subsubsection{Image Denoising}

Zheng et al. \cite{U-shaped-Mamba-Dehazing} proposes a classical U-shaped architecture based Mamba for image denoising, which can effectively capture local features and remote information. Its base block contains two consecutive convolutional blocks, and then the features are flattened and transposed to make them suitable to the input shapes for SSM. A two-branch Mamba structure is sequentially used to establish long-range dependencies in the sequence length and channel dimensions. Experiments on the RESIDE dataset, the low light enhancement task on the LOL dataset, the MIT-Adobe FiveK and the Deraining task on the Rain13K dataset show the effectiveness. FreqMamba \cite{FreqMamba} introduces frequency domain information into the Mamba model for image deraining. It takes into account the complex coupling of the raindrops to the background and the loss of important perceptual frequency information. Therefore, the models uses the Fourier transformation branch to provide frequency modeling and the wavelet packet transformation as a transition between the spatial Mamba branch and the Fourier Mamba branch. Finally, the outputs of the three parts are reconciled by concatenating the output features of the three branches and applying a $1 \times 1$ convolution operation.

\begin{table*}[t]
\caption{Comparison Between Vision Backbones with Mamba on ADE20K Dataset \cite{ADE20K} for Semantic Segmentation using UperNet \cite{UperNet}.}\label{Mamba_ADE20k}%
\setlength\tabcolsep{0.5pt}

\begin{tabular*}{\textwidth}{@{\extracolsep\fill}lccccc}
\toprule
Backbone & Image size & mIoU (SS) & mIoU (MS) & FLOPs (G) & \#Params (M). \\
\toprule
\textbf{CNN} \\
\toprule
ResNet-101 \cite{ResNet} & 38.2 & 58.8 & 41.4 & 34.7 & 55.7 \\
ConvNeXt-T \cite{ConvNeXt} & 44.7 & 65.8 & 48.3 & 40.1 & 63.3  \\
ConvNeXt-S \cite{ConvNeXt} & 45.4 & 67.9 & 50.0 & 41.8 & 65.2 \\
\toprule
\textbf{Transformer} \\
\toprule
Swin-T \cite{Swin} & 42.7 & 65.2 & 46.8 & 39.3 & 62.2 \\
Swin-S \cite{Swin} & 44.8 & 66.6 & 48.9 & 40.9 & 63.2 \\
PVTv2-B3 \cite{PVTv2} & 47.0 & 68.1 & 51.7 & 42.5 & 65.7 \\
\toprule
\textbf{Mamba}\\
\toprule
Vim-Ti \cite{VisionME} & $512^2$ & 41.0 & - & 13 & - \\
Vim-S \cite{VisionME} & $512^2$ & 44.9 & - & 46 & - \\
\midrule
VMamba-T \cite{VMamba} & $512^2$ & 47.3 & 48.3 & 55 & 939 \\
VMamba-S \cite{VMamba} & $512^2$ & 49.5 & 50.5 & 76 & 1037 \\
VMamba-B \cite{VMamba} & $512^2$ & 50.0 & 51.3 & 110 & 1167 \\
VMamba-S \cite{VMamba} & $640^2$ & 50.8 & 50.8 & 76 & 1620 \\
\midrule
LocalVim-T \cite{LocalMamba} & $512^2$ & 43.4 & 44.4 & 36 & 181 \\
LocalVim-S \cite{LocalMamba} & $512^2$ & 46.4 & 47.5 & 58 & 297 \\
LocalVMamba-T \cite{LocalMamba} & $512^2$ & 47.9 & 49.1 & 57 & 970 \\
LocalVMamba-S \cite{LocalMamba} & $512^2$ & 50.0 & 51.0 & 81 & 1095 \\
\midrule
EfficientVMamba-T \cite{EfficientVMamba} & $512^2$ & 38.9 & 39.3 & 14 & 230 \\
EfficientVMamba-S \cite{EfficientVMamba} & $512^2$ & 41.5 & 42.1 & 29 & 505 \\
EfficientVMamba-B \cite{EfficientVMamba} & $512^2$ & 46.5 & 47.3 & 62 & 930 \\
\midrule
SiMBA-S \cite{SiMBA} & $512^2$ & 49.0 & 49.6 & 62 & 1040 \\
\midrule
PlainMamba-Adapter-L1 \cite{PlainMamba} & $512^2$ & 44.1 & - & 35 & 174 \\
PlainMamba-Adapter-L2 \cite{PlainMamba} & $512^2$ & 46.8 & - & 55 & 285 \\
PlainMamba-Adapter-L3 \cite{PlainMamba} & $512^2$ & 49.1 & - & 81 & 419 \\
\bottomrule
\end{tabular*}
\end{table*}

\subsubsection{Image Restoration}

MambaIR \cite{MambaIR} uses VMamba architecture to explore the potential of Mamba in image restoration tasks. It feeds shallow feature into Residual State Space Blocks (RSSBs), which is a structure that adds a convolutional structure to VMamba's blocks for learning spatial local information. Furthermore, MambaIR introduces the Channel-Attention layer \cite{Channel-Attention-layer} to enhance the interaction between channels. Experiments are conducted on different image restoration tasks, including image super-resolution and real image denoising, and demonstrate the superiority of the method. Cheng et al. \cite{Activating_S_Resolution} propose to incorporate Vim into a MetaFormer-style block \cite{MetaFormer} for single-image super-resolution (SISR). To further expand the overall activation area, this paper introduces a complementary attention mechanism to process features in parallel with the Vim block. Experiments on various benchmark data demonstrate that it exhibits competitive and even superior performance compared to the state-of-the-art SISR methods while maintaining relatively low memory and computational overheads.

The application of U-shaped Mamba networks in image restoration has received attention. CU-Mamba \cite{CU-Mamba} applies spatial and channel SSM blocks to learn global context and channel features with only linear complexity. The model consists of three stages, each containing a CU-Mamba block and a down-sampling or up-sampling layer. In the internal structure, CU-Mamba contains a spatial SSM block followed by a channel SSM block. CU-Mamba follows the U-net setup, passes the feature outputs of the encoder stages to the decoder, and connects the encoder and decoder. In VmambaIR \cite{VmambaIR}, the encoder and decoder consist of an Omni Selective Scan (OSS) block. Specifically, the Omni Selective Scan Mechanism performs a pooling operation after scanning the spatial dimension in four directions. Then, it scans features in both forward and backward directions along the channel dimension. It also analyzes the ability of the Efficient Feed-Forward Network (EFFN) to perform layer normalization on features to mitigate pattern collapse and thus manage the information flow at each level more finely.

Retinexmamba \cite{Retinexmamba} introduces Mamba on the basis of Retinexformer \cite{Retinexformer} to build a novel Damage Restorer for low-light image enhancement output. Unlike Retinexformer, the encoder and decoder units of Retinexmamba's damage restorer are composed of the proposed novel Illumination Fusion State Space Model (IFSSM) which uses SS2D and a cross-attention mechanism to fuse illumination features and the input vector. 

\subsection{Mamba for 3D Vision}

\subsubsection{Point Cloud Analysis}
The point cloud is a collection of data consisting of a large number of points in 3D space, each with coordinates information and possibly other attributes such as color and intensity. Mamba exhibits powerful global modeling capabilities and linear computational complexity, which makes it attractive for point cloud analysis. However, due to Mamba's causality requirements and the disordered and irregular nature of point clouds, further modification of Mamba needs to be made in 3D point cloud tasks. 

PointMamba \cite{PointMamba} employs the lightweight PointNet \cite{PointNet} to embed point patches and generate point tokens. Subsequently, a simple but effective reordering strategy is applied to an order. It connects the point tokens along x, y and z axes based on the geometric coordinates of their clustering centers. However, it triples the length of the point tokens. 
Point Could Mamba \cite{PCM} further investigates the order. The point cloud data is fed into the Mamba by utilizing Consistent Traverse Serialisation(CTS) to the point cloud data. Specifically, CTS yields six variants by sequencing the coordinates to provide different views of the point cloud. A coding function is also designed to address the problem of large spatial distances between any two neighboring points. In addition, several learnable embeddings are added to the point cloud sequence at the beginning and end of the sequence to inform Mamba the order rules. 
Point Mamba \cite{Point_Mamba} is based on an octree sorting scheme. In contrast to PointMamba, the ordering strategy in this paper is used directly on the original input points instead of the clustering centers. The octree sorting scheme is able to sort the point cloud data stored in the octree nodes by shuffled keys to form a z-order based sequence. Compared to Transformer-based methods, Point Mamba achieves the state-of-the-art performance with 93.4\% accuracy and 75.7 mIOU on the ModelNet40 classification dataset and the ScanNet semantic segmentation dataset, respectively.
3DMambaComplete \cite{3DMambaComplete} uses plain Mamba blocks for input point cloud feature extraction and enhances the relationship between point features using Mamba's selectivity mechanism. Multi-head cross-attention and learnable offsets are introduced to predict the hyperpoints, which simultaneously avoids concentrating in a specific region.

\subsubsection{Hyperspectral Imaging Analysis}
The hyperspectral (HS) imaging system enables the simultaneous capture of spatial and spectral information by measuring the energy spectrum of each pixel. In HS data analysis, pixels and their corresponding spectral information must be accurately modeled. By analyzing these spectral signatures, hyperspectral imaging enables various applications such as material identification, classification and quantitative analysis. The emergence of Mamba has enhanced the utility of large-scale HS, and it is important to explore efficient Mamba frameworks for HS data.

Mamba-FETrack \cite{Mamba-FETrack} is a Frame-Event tracking framework that utilizes Vim to construct modality-specific backbone networks for extracting features of RGB frames and Event streams, respectively, and employs a FusionMamba block to facilitate the interaction and fusion of the features effectively. It achieves comparable performance on the FELT and FE108 datasets and shows huge advantages over ViT-based methods in terms of FLOPs and parameters.

\subsection{Mamba for Visual Generation}

For visual generation tasks, ZigMa \cite{ZigMa} incorporates Mamba structures into the diffusion model. Observing that various scanning strategies of previous Mamba approaches introduce additional parameters and GPU memory burdens, ZigMa spreads the complexity of Mamba to each layer. Its novelty lies in the assignment strategy of the eight scanning modes, as shown in Fig. \ref{scan}(h),  denoted as $S_j$ ($j\in [0, 7]$). The $\Omega_i = S_{i\%8}$ rule is used to assign these eight scanning modalities to the layers, where $\Omega$ is the arrangement of the tokens in layer $i$.

Motion Mamba \cite{Motion_Mamba} points out that the speed of classical diffusion model-based approaches in motion generation tasks is still affected by the quadratic complexity and leads to inefficiency. Therefore, Mamba is considered to address the complexity of long sequence generation. The Motion Mamba model is a motion generative system specifically designed to incorporate the diffusion of SSMs. It uses a denoising U-Net architecture to construct Hierarchical Temporal Mamba (HTM) and Bidirectional Spatial Mamba (BSM) in denoiser block based on SSM. Notably, this type of Mamba block employs multiple sub-SSMs in descending order of complexity, ensuring that the processing power is evenly distributed throughout the encoder-decoder architecture. Motion Mamba achieves up to 50\% improvement in FID on HumanML3D and KIT-ML datasets. SMCD \cite{SMCD} tackles motion style transfer task, which introduces motion style Mamba module for better temporal information extraction and long-term dependencies modeling.

Gamba \cite{Gamba} is an efficient feed-forward model that combines Gaussian splatting with Mamba for single-view 3D reconstruction. DINO v2 \cite{DINOv2} is used as the image tokenizer to obtain camera embedding, reference image tokens and a set of learnable 3D Gaussian Splatting embeddings. Subsequently, the Gamba block introduces cross-attention between them to facilitate context-dependent reasoning. In addition, this paper also scrutinizes the failure case of Gamba. For example, the authors found that it is difficult for Gamba to generate clear textures for occluded regions.

Gao et al. proposed Matten \cite{Matten} for Video Generation using Mamba-Attention architecture. The core of Matten lies in Global-Sequence Mamba Block with Spatial-Temporal Attention Interleaved, which uses spatiotemporal attention to model within a single frame, and uses bidirectional Mamba to model content between consecutive frames. Extensive experiments show that Matten not only embodies considerable computational efficiency, but can effectively capture global and local information existing in the video latent space.

\subsection{Discussion and Summary}
Various scanning mechanisms allow Mamba-based approaches to capture visual features sufficiently, and the extensibility and versatility of Mamba allow it to combine with other architectures and methodologies for wider vision challenges and applications from low-level to high-level, 2D to 3D, and discriminative to generative tasks.

\section{Mamba in Multi-Modal Learning Tasks}

Multiple different modalities (e.g., image, text, and audio) in performing specific tasks aimed at achieving more comprehensive and accurate data understanding and analysis. These tasks can cover multiple domains, such as natural language processing, computer vision and speech recognition. By integrating information from multiple modalities, the performance and robustness of the system can be improved. Mamba's global modeling capabilities have also shown potential in multi-modal tasks. This section introduces the performance of Mamba in multi-modal tasks from two aspects: heterologous stream and homologous stream. The heterologous stream is mainly concerned with the Mamba model of image interaction with other non image modalities, and the homologous stream is mainly concerned with the multi-modal fusion task of different types of images.

\subsection{Heterologous Stream}

ReMamba \cite{ReMamber} refers to image segmentation task. However, Mamba's inability to capture textual information and interactions across modalities is a significant challenge. To compensate for it, the authors propose a novel Mamba Twister, consisting of several layers of visual state space (VSS) and a twisting layer. By constructing a hybrid multi-modal feature cube, the twisting layer can fuse effectively textual and visual features in both channel and spatial dimensions. In addition, by comparing with the current popular Transformer-based methods, the authors points out it is necessary to consider the essential differences between Mamba and Transformer.

VL-Mamba \cite{VL-Mamba} uses a pre-trained Mamba as the backbone language model and the Vision Transformer (ViT) architecture as the visual encoder. In order to align non-causal visual data with causal 1D sequence data, the visual sequences are fed into a Multi-Modal Connector (MMC) with 2D vision selective scanning mechanisms. Then, this output vector is combined with a tokenized text query to generate the corresponding response. Although the approach differs slightly from several SOTA methods in some cases, with fewer parameters and limited training data, VL-Mamba achieves comparable performance compared to some models with more parameters. Cobra \cite{Cobra} concatenates visual features and text embedding as input into the Mamba backbone for multi-modal information fusion. The entire architecture requires only fine-tuning visual feature-related projectors and the Mamba backbone. It achieves performance comparable to that of LLaVA \cite{LLaVA} with about 43\% of the parameters on six commonly used VLM (vision-language model) benchmarks.

Mamba is also used in other multi-modal tasks such as Gesture Synthesis and Temporal video grounding. MambaTalk \cite{MambaTalk} uses audio and text sequences as inputs to co-direct four specialized Mamba models for gesture synthesis. SpikeMba \cite{SpikeMba} integrates Spiking Neural Networks and SSM to capture the fine-grained relationships between video and textual queries. Meanwhile, Spiking Neural Networks use a thresholding mechanism to identify salient objects efficiently, suppress noise and prevent the effects of sudden changes. Finally, a new multi-modal Relevant Mamba structure is proposed with a dual-input integration of the processed video and text features.

\subsection{Homologous Stream}

Sigma \cite{Sigma} is a Mamba model for multi-modal semantic segmentation. It introduces an attention mechanism in the VSS block to select relevant information from each modality. Specifically, Sigma exchanges the C-matrix during SSM computation of the two modalities to reconstruct the hidden state each other. Given the deficiency of Mamba in learning inter-channel information, the authors use the channel-attention operation for step-wise up-sampling in the decoder. In addition, residual connections of scaling parameters are introduced to enable the network to adjust the ratio between input and output adaptively. Sigma achieves excellent performance in various RGB-X semantic segmentation experiments. However, it only takes part of Mamba's ability to model longer sequences, and the four-direction scanning mechanism incurs high memory consumption. \cite{Fusion-Mamba-Detection} feeds images of two modalities into separate feature extraction networks. Then the two networks are connected using three Fusion-Mamba blocks (FMB) proposed for feature fusion. This module involves channel swapping for shallow feature fusion. Then the shallow fused features are projected into the hidden state space through a VSS block without gating to obtain $y_{R_i}$ and $y_{IR_i}$. Clearly, the gating parameters across branches reduce the differences between $y_{R_i}$ and $y_{IR_i}$ to achieve feature enhancement.

\subsection{Discussion and Summary}
The architecture of Mamba enables the model to flexibly adapt to different data types as well as task requirements. Its powerful scalable sequences modeling capability facilitates cross-modal interaction and fusion in multi-modal learning.

\section{Mamba in Vertical-Domain Tasks}

Due to the properties of linear complexity and higher inference throughput than other architectures, Mamba has attracted much interest in a variety of downstream tasks. This section provides an overview of Mamba in vertical-domain visual applications, including processing, analysis and understanding of remote sensing and medical images.

\subsection{Mamba for Remote Sensing Image Modeling}
\subsubsection{Remote Sensing Image Processing}
Considering the effectiveness of the Mamba model in global information modeling, Pan-Mamba \cite{Pan-Mamba} first attempts to introduce Mamba to the pan-sharpening task. Its network architecture consists of three key components: the Mamba block for extracting features and modeling long-range dependencies, the channel swapping Mamba block for implementing lightweight feature interactions across modalities and initiating a correlation between them, and the cross-modality Mamba block for facilitating the learning of complementary features while suppressing redundant features using a gating mechanism. HSIDMamba \cite{HSIDMamba} comprises multiple hyperspectral continuous scan blocks, incorporating scale residual, spectral attention mechanisms, and a bidirectional continuous scanning mechanism meticulously tailored to the nuances of hyperspectral images to capture spatial-spectral dependencies for hyperspectral image denoising effectively.

\subsubsection{Remote Sensing Image Classification}
RSMamba \cite{RSMamba} is designed based on SSM and Mamba to enhance the whole-image comprehension for remote sensing image classification while exploiting a position-sensitive dynamic multi-path activation mechanism to extend Mamba for 2D non-causal data. SpectralMamba \cite{SpectralMamba} is a lightweight state space model for hyperspectral image classification that utilizes a piece-wise sequential scanning strategy to leverage the reflectance characteristics of different types of ground objects, as well as a gated spatial-spectral merging strategy to encode the spatial regularity and spectral peculiarity adequately. SS-Mamba \cite{SS-Mamba} stacks multiple spectral-spatial Mamba blocks to construct a hyperspectral image classification framework, where each spectral-spatial Mamba block processes the spatial token and spectral token separately, and ultimately modulates the spatial and spectral tokens with the information of the central region of the HIS sample for spectral-spatial information fusion. S2Mamba \cite{S2Mamba} contains a patch cross scanning mechanism and a bi-direction spectral scanning mechanism, where the former captures spatial contextual features through the interaction of each pixel with its neighboring pixels, and the latter captures spectral contextual features by a bi-directional interaction between each band, and introduces a spatial-spectral mixture gate for adaptive fusion to boost hyperspectral image classification.

\subsubsection{Remote Sensing Image Change Detection}
ChangeMamba \cite{ChangeMamba} explores the potential of the Mamba architecture for remote sensing change detection tasks, designs different frameworks for binary change detection, semantic change detection, and builds damage assessment, respectively. The encoder is designed with Visual Mamba architecture to learn sufficient global spatial contextual information, and for the change decoder, sequential, cross and parallel modeling mechanisms are further exploited, through which the spatio-temporal interaction of multi-temporal features can be achieved. Therefore, more accurate change detection results are obtained. RSCama \cite{RSCama} tackles remote sensing image change caption, and its core components are spatial difference-guided SSM and temporal traveling SSM, where the former enhances change perception in spatial features and the latter facilitates bi-temporal interactions through token-wise cross-scanning.

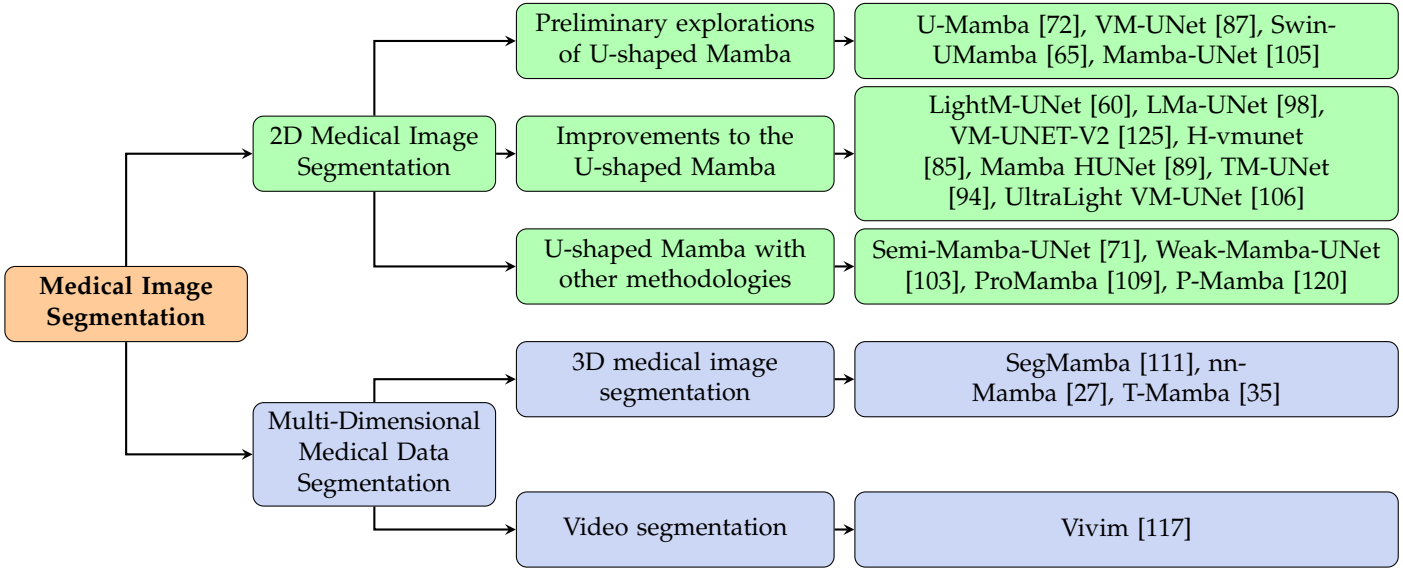
\begin{figure*}[t]
\centering
\tikzstyle{begin} = [rectangle, rounded corners, minimum width=3cm, minimum height=1cm, text centered, text width=3cm, draw=black, fill=orange!40]
\tikzstyle{2D} = [rectangle, rounded corners, minimum width=3cm, minimum height=1cm, text centered, text width=3cm, draw=black, fill=green!30]
\tikzstyle{2D_middle} = [rectangle, rounded corners, minimum width=3cm, minimum height=1cm, text centered, text width=4cm, draw=black, fill=green!30]
\tikzstyle{3D} = [rectangle, rounded corners, minimum width=3cm, minimum height=1cm, text centered, text width=3cm, draw=black, fill=green!20!blue!20]
\tikzstyle{3D_middle} = [rectangle, rounded corners, minimum width=3cm, minimum height=1cm, text centered, text width=4cm, draw=black, fill=green!20!blue!20]
\tikzstyle{2D_end} = [rectangle, rounded corners, minimum width=3cm, minimum height=1cm, text centered, text width=7cm, draw=black, fill=green!30]
\tikzstyle{3D_end} = [rectangle, rounded corners, minimum width=3cm, minimum height=1cm, text centered, text width=7cm, draw=black, fill=green!20!blue!20]
\tikzstyle{arrow} = [thick,->,>=stealth]
\begin{tikzpicture}[node distance=2cm]
\node (start) [begin] {\textbf{Medical Image Segmentation}};
\node (2D) [2D, right of=start, xshift=1.3cm, yshift=2cm] {2D Medical Image Segmentation};
\node (3D) [3D, right of=start, xshift=1.3cm, yshift=-2cm] {Multi-Dimensional Medical Data Segmentation};
\node (2D_middle1) [2D_middle, right of=2D, xshift=2cm, yshift=1.5cm] {Preliminary explorations of U-shaped Mamba};
\node (2D_middle2) [2D_middle, right of=2D, xshift=2cm] {Improvements to the U-shaped Mamba};
\node (2D_middle3) [2D_middle, right of=2D, xshift=2cm, yshift=-1.5cm] {U-shaped Mamba with other methodologies};
\node (3D_middle1) [3D_middle, right of=3D, xshift=2cm, yshift=1cm] {3D medical image segmentation};
\node (3D_middle2) [3D_middle, right of=3D, xshift=2cm, yshift=-1cm] {Video segmentation};
\node (2D_end1) [2D_end, right of=2D_middle1, xshift=4cm] {U-Mamba \cite{U_Mamba}, VM-UNet \cite{VM-UNet}, Swin-UMamba \cite{Swin-UMamba}, Mamba-UNet \cite{Mamba-UNet}};
\node (2D_end2) [2D_end, right of=2D_middle2, xshift=4cm] {LightM-UNet \cite{LightM-UNet}, LMa-UNet \cite{LMa-UNet}, VM-UNET-V2 \cite{VM-UNET-V2}, H-vmunet \cite{H-vmunet}, Mamba HUNet \cite{Mamba-HUNet}, TM-UNet \cite{TM-UNet}, UltraLight VM-UNet \cite{UltraLight_VM-UNet}};
\node (2D_end3) [2D_end, right of=2D_middle3, xshift=4cm] {Semi-Mamba-UNet \cite{Semi-Mamba-UNet}, Weak-Mamba-UNet \cite{Weak-Mamba-UNet}, ProMamba \cite{ProMamba}, P-Mamba \cite{P-Mamba}};
\node (3D_end1) [3D_end, right of=3D_middle1, xshift=4cm] {SegMamba \cite{SegMamba}, nnMamba \cite{nnMamba}, T-Mamba \cite{T-Mamba}};
\node (3D_end2) [3D_end, right of=3D_middle2, xshift=4cm] {Vivim \cite{Vivim}};
\draw [arrow] (start) |- (2D);
\draw [arrow] (start) |- (3D);
\draw [arrow] (2D) |- (2D_middle1);
\draw [arrow] (2D) -- (2D_middle2);
\draw [arrow] (2D) |- (2D_middle3);
\draw [arrow] (3D) |- (3D_middle1);
\draw [arrow] (3D) |- (3D_middle2);
\draw [arrow] (2D_middle1) -- (2D_end1);
\draw [arrow] (2D_middle2) -- (2D_end2);
\draw [arrow] (2D_middle3) -- (2D_end3);
\draw [arrow] (3D_middle1) -- (3D_end1);
\draw [arrow] (3D_middle2) -- (3D_end2);
\end{tikzpicture}
\caption{Taxonomy of studies focusing on Mamba in medical image segmentation.}
\label{med_seg_Taxonomy}
\end{figure*}

\subsubsection{Remote Sensing Image Segmentation}
Samba \cite{Samba} is a specialized framework for high-resolution remote sensing image segmentation based on Mamba, demonstrating unparalleled performance on a range of public datasets and outperforming the state-of-the-art CNN and Transformer-based methods. RS3Mamba \cite{RS3Mamba} is a dual-branch network for remote sensing image semantic segmentation, where the auxiliary branch is designed based on visual state space blocks for providing additional global information to the CNN-based main branch. In addition, a collaborative completion module is designed to enhance and fuse cross-branch features from both local and global perspectives. RS-Mamba \cite{RS-Mamba} focuses on dense prediction tasks of very-high-resolution remote sensing images and achieves efficient capture of global contextual information. In addition, an omnidirectional selective scan module is designed to model the image contextual information in different directions globally, considering the spatial direction distribution characteristics of remote sensing images.

\subsubsection{Remote Sensing Image Fusion}
FusionMamba \cite{FusionMamba_Efficient} is dedicated to image fusion, incorporating Mamba blocks into two U-shaped networks to extract spatial and spectral features independently and hierarchically, and designing a fusion module to simulate the gradual injection of spatial information into the spectral feature maps to achieve comprehensive information fusion. LE-Mamba \cite{State_Sharing_Fusion} is a visual Mamba-based multi-scale network designed for image fusion, equipped with local-enhanced vision Mamba blocks to represent and combine local and global spatial information. The state sharing technique aims to reduce information loss and enhance spatial details.

\subsection{Mamba for Medical Image Modeling}
\subsubsection{Medical Image Segmentation}

The related works in medical image segmentation can be categorized into 2D medical image segmentation and multi-dimensional medical data segmentation. Consider there has been a number of Mamba models for medical image segmentation, we present a detailed taxonomy in Fig.~\ref{med_seg_Taxonomy}.

\emph{Preliminary explorations of U-shaped Mamba}. U-Mamba \cite{U_Mamba} is a general-purpose medical image segmentation architecture that integrates the merits of CNN for extracting local features and SSM for capturing global information and outperforms CNN-based and Transformer-based architectures. VM-UNet \cite{VM-UNet}, the first pure SSM-based medical image segmentation model with an asymmetrical encoder-decoder structure, demonstrates superior performance on ISIC17, ISIC18 and Synapse datasets. Swin-UMamba \cite{Swin-UMamba} is a Mamba-based U-shaped model and demonstrates the importance of ImageNet-based pretraining in improving the performance of Mamba-based models. Swin-UMamba$^{\dagger}$ is an enhanced variant that replaces the CNN-based decoder with a Mamba-based decoder, and achieves competitive results while using fewer network parameters and imposing a lower computational cost than Swin-UMamba. Both architectures demonstrate performance beyond CNN-based, Transformer-based and Mamba-based models on AbdomenMRI, Endoscopy and Microscopy datasets. Mamba-UNet \cite{Mamba-UNet} is also a pure SSM-based medical image segmentation model, and unlike VM-UNet, the encoder and decoder of Mamba-UNet are mirrored, and the bottleneck consists of two visual Mamba blocks. Mamba-UNet outperforms classical U-shaped CNN and Transformer architectures on ACDC MRI Cardiac and Synapse CT Abdomen segmentation datasets.

\emph{Improvements to the U-shaped Mamba}. LightM-UNet \cite{LightM-UNet} employs Mamba as a lightweight strategy of UNet to alleviate the demand for computational resources in real medical settings and further enhances the ability of SSM to model long-range spatial dependencies by utilizing residual connectivity and adjustment factors. Experimental validation on the LiTs dataset and the Montgomery\&Shenzhen dataset demonstrates that LightM-UNet outperforms competing architectures with fewer network parameters and lower computational overheads. LMa-UNet \cite{LMa-UNet} achieves large spatial modeling by assigning large windows to SSM modules, designs hierarchical Mamba blocks for location-aware sequence modeling and bidirectional Mamba blocks for pixel- and patch-level feature modeling, and possesses excellent local spatial modeling and efficient global modeling capabilities. Following the framework of UNetV2, VM-UNetV2 \cite{VM-UNET-V2} introduces VSS blocks to capture contextual information and a Semantics and Detail Infusion module to facilitate the interaction and fusion of low-level and high-level features. VM-UNetV2 achieves competitive performance on the ISIC17, ISIC18, CVC-300, CVC-ClinicDB, Kvasir, CVC ColonDB and ETIS AribPolypDB datasets by initializing the encoder with VMamba pretrained weights and employing a deep supervision mechanism. H-vmunet \cite{H-vmunet} exploits a high-order 2D-selective scan to progressively reduce the introduction of redundant information, while keeping a superior global receptive field and boosting the learning of local feature information. Mamba-HUNet \cite{Mamba-HUNet} is a multi-scale hierarchical up-sampling network that efficiently captures local features as well as long-range dependencies in medical images by exploiting the linear scaling advantage of Mamba and the global contextual understanding capability of HUNet. TM-UNet \cite{TM-UNet} uses the residual connection to enhance the ability of VSS blocks to extract local and global features and uses Triplet Attention-inspired Triplet-SSM as the bottleneck to integrate spatial and channel features. Wu et al. \cite{UltraLight_VM-UNet} delve into the key factors affecting the Mamba parameters and propose a parallel vision Mamba method, i.e.,  UltraLight VM-UNet for processing deep features with low computational complexity. It achieves competitive performance for skin lesion segmentation on three public datasets, including ISIC2017, ISIC2018 and PH2, while significantly reducing parameters.

\emph{U-shaped Mamba with other methodologies}. Considering the computational burden faced by CNNs and ViTs in dealing with long-range dependencies, as well as the high cost and even unavailability of expert annotations, Wang et al. \cite{Semi-Mamba-UNet} proposed Semi-Mamba-UNet, which integrates a visual Mamba-based UNet architecture with a conventional UNet to jointly generate pseudo-labels and cross-supervise each other in a semi-supervised learning framework, combined with a self-supervised contrastive learning strategy to boost feature learning. Weak-Mamba-UNet \cite{Weak-Mamba-UNet} is a weakly supervised learning framework for scribble-based medical image segmentation that combines CNN-based UNet, Swin Transformer-based SwinUNet and VMamba-based Mamba-UNet to facilitate iterative learning and refinement across networks using pseudo labels in a collaborative as well as a cross-supervised manner. ProMamba \cite{ProMamba} is the first attempt to introduce prompt and Vision Mamba into the polyp segmentation task, where Vision Mamba features powerful feature extraction capability, while box prompt boosts the generalization ability. P-Mamba \cite{P-Mamba} is specifically designed to relieve the challenges of the lack of efficiency as well as the background noise interference faced by pediatric echocardiographic left ventricular segmentation, with DWT-based Perona-Malik diffusion blocks for noise suppression and local feature extraction as well as Vision Mamba for efficient global dependency modeling.

\emph{Multi-Dimensional Medical Data Segmentation}. SegMamba \cite{SegMamba} is the first general 3D medical image segmentation framework developed based on Mamba, which integrates a U-shaped structure with Mamba to model global features at multiple scales. It exploits a gated spatial convolution module to boost the spatial feature representation before each tri-orientated Mamba module, which models 3D features from three directions. nnMamba \cite{nnMamba} integrates the strengths of CNNs and SSMs to model local and global features using the Mamba-In-Convolution with Channel-Spatial Siamese input module and can be used as a backbone for a variety of 3D medical image tasks, including 3D image segmentation, classification and landmark detection. T-Mamba \cite{T-Mamba} is proposed for tooth CBCT segmentation, which is the first work to introduce frequency features into visual Mamba. It boosts spatial position preservation and frequency-domain feature enhancement by integrating shared position coding and frequency-based features into the visual Mamba and designs a gated selection unit adaptively integrating two spatial-domain features and one frequency-domain feature. Vivim \cite{Vivim} is the first work to incorporate SSM into the task of medical video object segmentation, employing temporal Mamba blocks to efficiently compress long-term spatiotemporal representations into sequences at different scales, and introduces a boundary-aware constraint to enhance the predicted boundary structure. The core of the temporal Mamba block lies in the structured state space models with spatiotemporal selective scan, which explicitly consider single-frame spatial coherence and cross-frame coherence.

\begin{table*}[t]
\caption{Experimental results on organ segmentation in abdomen MRI scans, instruments segmentation in endoscopy images, and cell segmentation in microscopy. Abdomen MRI: The dataset was from the MICCAI 2022 AMOS Challenge \cite{Abdomen_MRI}. Endoscopy images: The dataset was from the MICCAI 2017 EndoVis Challenge \cite{Endoscopy_image}. Microscopy images:
The datasets was from the NeurIPS 2022 Cell Segmentation Challenge \cite{Microscopy_images}. $^{*}$ represents the best result for the corresponding model.}\label{Mamba_medicine_2d_MR}
\small
\begin{tabular*}{\textwidth}{@{\extracolsep\fill}lccccc}
\toprule%
& \multicolumn{2}{c@{}}{Organ in Abdomen MRI} & \multicolumn{2}{c@{}}{Instruments in Endoscopy} & {Cell in Microscopy} \\\cmidrule{2-3}\cmidrule{4-5}\cmidrule{6-6}%
Methods  & DSC & NSD & DSC & NSD & F1 \\
\toprule
\textbf{CNN} \\
\toprule
nnU-Net \cite{nnU-Net} & 0.7450 & 0.8153 & 0.6264 & 0.6412 & 0.5383 \\
SegResNet \cite{SegResNet} & 0.7317 & 0.8034 & 0.5820 & 0.5968 & 0.5411  \\
\toprule
\textbf{Transformer} \\
\toprule
UNTER \cite{UNETR} & 0.5474 & 0.6309 & 0.5071 & 0.5168 & 0.4357 \\
SwinUNETR \cite{Swin_UNETR} & 0.7028 & 0.7669 & 0.5528 & 0.5683 & 0.3967 \\
nnFormer \cite{nnFormer} & 0.7279 & 0.7963 & 0.6135 & 0.6228 & 0.5332 \\

\toprule
\textbf{Mamba} \\
\toprule
U-Mamba$^*$ \cite{U_Mamba}  & 0.7625 & 0.8327 & 0.6540 & 0.6692 & 0.5607 \\
Swin-UMamba$^*$ \cite{Swin-UMamba}  & 0.7760 & 0.8421 & 0.6783 & 0.6933 & 0.5982 \\
LMa-UNet \cite{LMa-UNet} & 0.7735 & 0.8380 & - & - & - \\

\bottomrule
\end{tabular*}
\end{table*}

\begin{table*}[t]
\caption{Experimental results on ISIC17 \cite{ISIC2017} and ISIC18 \cite{ISIC2018}. The results are in percentage (\%). Note that Acc., Spe. and Sen. mean accuracy, specificity and sensitivity, respectively.}\label{Mamba_medicine_2d_ISIC}
\small
\begin{tabular*}{\textwidth}{@{\extracolsep\fill}lcccccccccc}
\toprule%
& \multicolumn{5}{c@{}}{ISIC17} & \multicolumn{5}{c@{}}{ISIC18}\\ \cmidrule{2-6}\cmidrule{7-11}
Methods  & mIoU & DSC & Acc. & Spe. & Sen. & mIoU & DSC & Acc. & Spe. & Sen. \\
\toprule
UNet \cite{U-Net}
& 76.98 & 86.99 & 95.65 & 97.43 & 86.82
& 77.86 & 87.55 & 94.05 & 96.69 & 85.86 \\
UTNetV2 \cite{UTNetV2}
& 77.35 & 87.23 & 95.84 & 98.05 & 84.85
& 78.97 & 88.25 & 94.32 & 96.48 & 87.60 \\
TransFuse \cite{TransFuse}
& 79.21 & 88.40 & 96.17 & 97.68 & 87.14
& 80.63 & 89.27 & 94.66 & 95.74 & 91.28 \\
MALUNet \cite{MALUNet}
& 78.78 & 88.13 & 96.18 & 98.47 & 84.78
& 80.25 & 89.04 & 94.62 & 96.19 & 89.74 \\
\toprule
VM-UNet \cite{VM-UNet}
& 80.23 & 89.03 & 96.29 & 97.58 & 89.90
& 81.35 & 89.71 & 94.91 & 96.13 & 91.12 \\
VM-UNET-V2 \cite{VM-UNET-V2}
& 82.34 & 90.31 & 96.70 & 97.67 & 91.89
& 81.37 & 89.73 & 95.06 & 97.13 & 88.64 \\
H-vmunet \cite{H-vmunet}
& - & 91.72 & 90.56 & 98.31 & 96.80
& - & - & - & - & - \\
TM-UNet \cite{TM-UNet}
& 80.51 & 89.20 & 96.46 & 98.28 & 87.37
& 81.55 & 89.84 & 95.08 & 96.68 & 89.98 \\
UltraLight VM-UNet \cite{UltraLight_VM-UNet}
& - & 90.91 & 96.46 & 97.90 & 90.53
& - & 89.40 & 95.58 & 97.81 & 86.80 \\
\bottomrule
\end{tabular*}
\end{table*}

\begin{table*}[t]
\caption{Experimental results on MRI Cardiac Test Set \cite{MRICardiac}. $^{*}$ represents the best result for the corresponding model.}\label{Mamba_medicine_MRI_Cardiac}
\small
\begin{tabular*}{\textwidth}{@{\extracolsep\fill}lccccccc}
\toprule%
Methods  & Dice & Acc. & Pre. & Sen. & Spe. & HD & ASD \\
\toprule
UNet \cite{U-Net}
& 0.9248 & 0.9969 & 0.9157 & 0.9364 & 0.9883 & 2.7655 & 0.8180\\
Swin-UNet \cite{Swin-Unet}
& 0.9188 & 0.9968 & 0.9151 & 0.9231 & 0.9857 & 3.1817 & 0.9932 \\
\toprule
Mamba-UNet \cite{Mamba-UNet}
& 0.9281 & 0.9972 & 0.9275 & 0.9289 & 0.9859 & 2.4645 & 0.7677 \\
Semi-Mamba-UNet$^*$ \cite{Semi-Mamba-UNet}
& 0.9114 & 0.9964 & 0.9088 & 0.9146 & 0.9821 & 3.9124 & 1.1698 \\
Weak-Mamba-UNet \cite{Weak-Mamba-UNet}
& 0.9197 & 0.9963 & 0.9095 & 0.9309 & 0.9920 & 3.9597 & 0.8810 \\

\bottomrule
\end{tabular*}
\end{table*}

\subsubsection{Pathological Diagnosis}

Yue et al. \cite{MedMamba} proposed MedMamba for medical image classification, which utilizes a designed Conv-SSM module that allows the model to capture long-range dependencies while extracting local features efficiently. MedMamba aims to construct a new baseline for the task of medical image classification and demonstrates considerable competitiveness on multiple datasets. MamMIL \cite{MamMIL} introduces Mamba to multiple instance learning (MIL) and achieves efficient whole slide image (WSI) classification. It introduces a bidirectional state space model as well as a 2D content-aware block based on pyramid-structured convolutions to learn bidirectional instance dependencies with 2D spatial relations. MamMIL demonstrates advanced classification performance and consumes less GPU memory than Transformer-based methods on Camelyon16 and BRACS datasets. Considering the limitations of existing MIL approaches in facilitating comprehensive and effective interactions between instances, Yang et al. \cite{MambaMIL} proposed MambaMIL with Sequence Reordering Mamba, which can be aware of the order and distribution of instances. As the core component to efficiently capture more discriminative features, it reduces the risk of overfitting and decreases the computational overhead. Benefiting from long sequence modeling, MambaMIL demonstrates superior performance on nine public datasets for both survival prediction and cancer subtyping tasks. CMViM \cite{CMViM} is the first efficient representation learning method for 3D multi-modal data designed for Alzheimer's disease classification. It combines vision Mamba with the mask autoencoder to efficiently model the intrinsic long-range dependencies of 3D medical data, and exploits intra-modal contrast learning and inter-modal contrast learning for modeling discriminative features of the same modality as well as aligning cross-modal representations, respectively. SurvMamba \cite{SurvMamba} explores multi-grained and multi-modal interactions for survival prediction. It utilizes the hierarchical interaction Mamba module to facilitate efficient interactions of intra-modal features at different levels of granularity and the Interaction Fusion Mamba module to facilitate interaction and fusion of inter-modal features across different levels.

\subsubsection{Deformable Image Registration}

MambaMorph \cite{MambaMorph} processes the medical MR-CT deformable alignment task, utilizing an alignment module that incorporates Mamba blocks to achieve long-range spatial relationship modeling and reduce the computational burden and a fine-grained feature extractor with a U-shaped structure to achieve high-dimensional feature learning. VMambaMorph \cite{VMambaMorph} further redesigns 2D image-based VSS blocks for 3D feature processing and employs a recursive registration framework to achieve coarse-to-fine alignment, demonstrating performance beyond MambaMorph.

\subsubsection{Medical Image Reconstruction}

FDVM-Net \cite{FDVM-Net} achieves high-quality endoscopic exposure correction through frequency domain reconstruction. It employs the Mamba and convolutional blocks to design the basic unit for extracting local features and modeling long-range dependencies, based on which a dual-path network is designed to process the phase and amplitude information of images separately, with a frequency domain cross-attention module to boost performance. Huang et al. \cite{MambaMIR} developed Mamba-based MambaMIR and its GAN variant MambaMIR-GAN for medical image reconstruction and uncertainty estimation. In addition, an arbitrary-mask mechanism is applied to adapt Mamba to image reconstruction effectively and introduce randomness for subsequent uncertainty estimation. FusionMamba \cite{FusionMamba_Dynamic} is an efficient Mamba model for image fusion that integrates the visual state space model with dynamic convolution and channel attention to dynamically enhance local features while reducing channel redundancy, and employs a dynamic feature fusion module to enhance the detailed texture information and differential information of the source image as well as to boost a better information interaction between modalities. MambaDFuse \cite{MambaDFuse} is also a model tailored for image fusion, which utilizes a dual-level feature extractor designed by CNN and Mamba blocks to capture long-range features from single-modality images, and introduces a dual-phase feature fusion module to obtain fused features with complementary information of different modalities via channel exchange and enhanced multi-modal Mamba blocks.

\subsubsection{Other Medical Tasks}
Fu et al. \cite{MD-Dose} develops a Mamba-based diffusion model MD-Dose for radiation dose prediction, which consists of a Mamba-structured noise predictor integrated with a Mamba encoder to extract structural information. Experimental results on a private dataset show that MD-Dose achieves superior performance while surpassing other diffusion model methods in inference speed. Zhang et al. \cite{Dual-Camera_Tracker} applies Mamba to object tracking and motion processing for the first time and designs a Mamba-based motion-guided prediction head to construct motion tokens from long-range temporal dependencies to explore the latent information from historical motion sufficiently.

\subsection{Discussion and Summary}
The linear complexity and global modeling capabilities of Mamba make it naturally suitable for processing high-resolution remote sensing images and medical images. It shows significant advantages in computational efficiency, GPU memory consumption, inference speed, etc., greatly promoting the deployment and implementation of deep learning models in practical applications.

\section{Conclusion, Challenge and Outlook}
Mamba is receiving an unprecedented attention in the field of computer vision and its vertical domains as a new alternative of network architecture. Mamba has shown a great potential in both alternative and composite architectures due to its unique features from popular convolutional neural network and Transformer architectures. This article presents a comprehensive review of Mamba and its variants deployed to various visual tasks, including general vision such as high/mid-level vision, low-level vision, 3D Vision and visual generation, vision-language multi-modal learning, and vertical domains such as remote sensing intelligence and medical intelligence. Nevertheless, Mamba is standing in an early stage, 
and there is still much room for improvement when compared to those relatively mature Transformer-based models. 
We offer several potential challenges and possible directions for future research to further advance Mamba.

\subsection{New Scanning Mechanism}
Mamba must obtain the state of the subsequent time point based on the previous time point and hidden state and result in causal properties. Visual data has non-causal properties and also has certain spatial relationships. It is necessary to consider how to remove the spurious association and retain the necessary image structure information when deploying Mamba. 
The earliest visual Mamba backbones formulated from the perspective of scanning operations, such as Vim and VMamba, proposed bidirectional scanning mechanisms and cross-scanning mechanisms, respectively. After that, different scanning schemes are proposed to bridge the gap in the input data and adapt it to downstream tasks. Therefore, it is necessary to design a reasonable scanning mechanism to improve Mamba's performance in visual tasks.

\subsection{Synergistic Hybrid Architecture}
Mamba has advantages over Transformers but there are also some essential drawbacks, such as insufficient interactions between tokens due to the lack of attention mechanism, which adversely affects the capture of more comprehensive and detailed information. Therefore, building hybrid models incorporating Mamba would be a promising direction to reduce this inherent shortcoming effectively. Prior to Mamba, \cite{Efficient_Movie} combines the advantages of structured state-space sequences (S4) and self-attention layers to capture short-range intra-camera dependencies and aggregate long-range inter-camera cues. Nevertheless, how to combine Mamba with Transformer for vision tasks is still a challenge, e.g., \cite{ReMamber} experimentally show that the combination of the VSS block and cross-attention mechanism does not work well for image segmentation. Hybrid models need to consider the fundamental differences between Mamba and other architectures. For example, Mamba models are strictly sequential in their predictions and usually do not modeling tokens not appeared in the scan, whereas the attention mechanism treats all tokens equally. This essential difference in the way the sequences are modeled easily makes both architectures conflict, 
and therefore bridging the gap is pretty important. 

\subsection{Following the Law of Scale}
Although Mamba has attracted attention for its remarkable computational efficiency, improving model capacity is still essential in the current era of large models following the scaling law. Currently, large-scale model with Mamba architecture is not yet appeared. Large-scale models incorporating the advantages of Mamba may be more powerful in long-term sequence modeling and another competitive visual foundation model. Moreover, various Mamba tuning methods will emerge as the model scale increases.

\subsection{Integration with Other Methodologies}
Mamba, as a base architecture, can also serve other methodologies, such as multi-modal information processing, diffusion models, domain generalization, visual-language models, etc. Following Mamba principle, how to work with these methods, still needs to be further explored.


%

\appendices


%
%

\ifCLASSOPTIONcaptionsoff
  \newpage
\fi



%

\bibliographystyle{IEEEtranS}
\bibliography{sn-bibliography}
%

%
\begin{IEEEbiography}[{\includegraphics[width=1.2in,height=1.2in,clip,keepaspectratio]{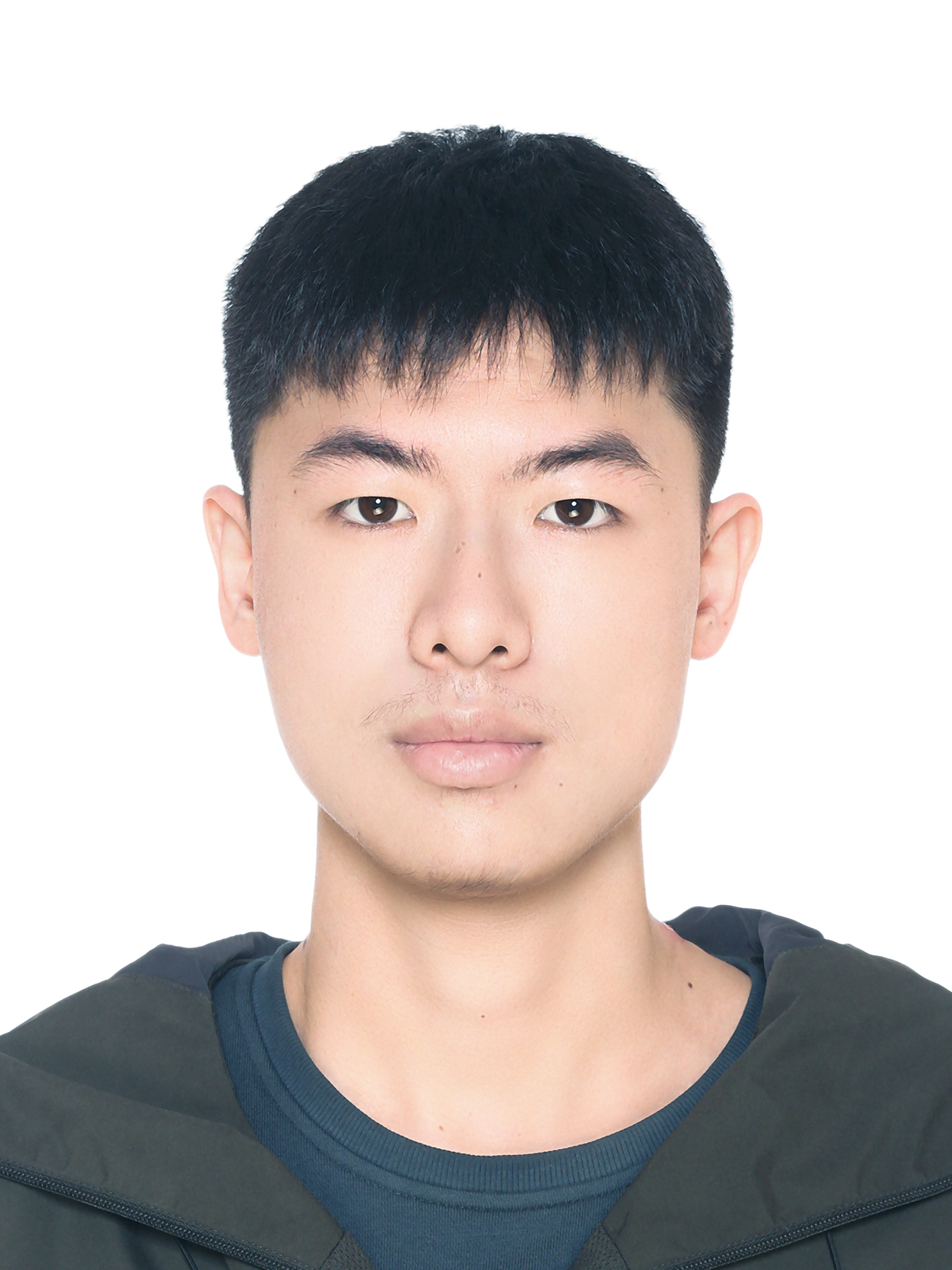}}]{Xiao Liu}
received the B.S. degree from the Chongqing University, China, in 2023. He is currently pursuing the M.S. degree at Chongqing University, China.

His research interests include deep learning, computer vision and semantic segmentation.
\end{IEEEbiography}

\begin{IEEEbiography}[{\includegraphics[width=1.2in,height=1.2in,clip,keepaspectratio]{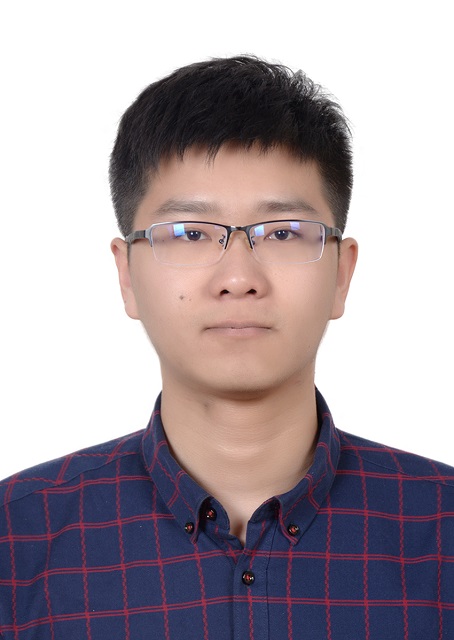}}]{Chenxu Zhang}
received the B.S. degree from the North China Electric Power University, China, in 2019, and the M.S. degree from Chongqing University, China, in 2022. He is currently pursuing the Ph.D degree at Chongqing University, China.

His research interests include deep learning, computer vision and semantic segmentation.
\end{IEEEbiography}

\begin{IEEEbiography}[{\includegraphics[width=1.2in,height=1.2in,clip,keepaspectratio]{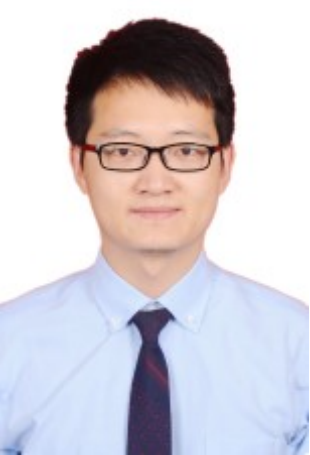}}]{Lei Zhang}
(M'14-SM'18) received his Ph.D degree in Circuits and Systems from the College of Communication Engineering, Chongqing University, Chongqing, China, in 2013. He worked as a Post-Doctoral Fellow with The Hong Kong Polytechnic University, Hong Kong, from 2013 to 2015. He is currently a Professor with Chongqing University. He has authored more than 100 scientific papers in top journals and conferences, including IEEE TPAMI, IJCV, CVPR, ICCV, ECCV, ICML, etc. He is on the Editorial Boards of several journals, such as IEEE Transactions on Instrumentation and Measurement, Neural Networks (Elsevier), etc. 
His current research interests include computer vision, deep learning and transfer learning. He is a Senior Member of IEEE.
\end{IEEEbiography}

\end{document}